\documentclass[11pt]{article}
\usepackage{Frontiers2021}
\colingfinalcopy
\usepackage{times}
\usepackage{url}
\usepackage{latexsym}
\usepackage{booktabs}
\usepackage{amsmath}
\usepackage{graphicx}
\usepackage{tabularx}
\usepackage{comment}
\usepackage{todonotes}
\usepackage{pgfplots}
\pgfplotsset{compat=1.14}
\usepackage{pgfplotstable}

\definecolor{aqua}{rgb}{0.0, 1.0, 1.0}

\newlength\graphheight
\newlength\graphwidth
\pgfplotsset {
graph/.style={
height=\graphheight,
width=\graphwidth,
legend style={
    draw=none,
    fill=none,
    font=\footnotesize,
},
ticklabel style={font=\scriptsize, /pgf/number format/fixed},
scaled y ticks = false,
scaled x ticks = false,
xlabel near ticks,
ylabel near ticks,
},
graphright/.style={
graph,
yticklabel pos=right,
},
}

\title{\textsc{Synthetic4Health}: Generating Annotated Synthetic Clinical Letters}

\author{
           Libo Ren, Samuel Belkadi, Lifeng Han$^{*}$ \\ \textbf{Warren Del-Pinto},
\and \textbf{Goran Nenadic}
\\
         The University of Manchester, UK \\ 
         $^{*}$ \textit{corresponding author}
         \\ {\tt lifeng.han, warren.del-pinto, g.nenadic@manchester.ac.uk} 
         \\
         {\tt 
        renlibo994, belkadisamuel@gmail.com 
        }  \\             }
\date{}

\begin{document}
\maketitle

\begin{abstract}
Since clinical letters contain sensitive information, clinical-related datasets can not be widely applied in model training, medical research, and teaching. This work aims to generate reliable, various, and de-identified synthetic clinical letters. To achieve this goal, we explored different pre-trained language models (PLMs) for masking and generating text.
After that, we worked on Bio\_ClinicalBERT, a high-performing model, and experimented with different masking strategies. Both qualitative and quantitative methods were used for evaluation. Additionally, a downstream task, Named Entity Recognition (NER), was also implemented to assess the usability of these synthetic letters. 

The results indicate that 1) encoder-only models outperform encoder-decoder models. 2) Among encoder-only models, those trained on general corpora perform comparably to those trained on clinical data when clinical information is preserved. 3) Additionally, preserving clinical entities and document structure better aligns with our objectives than simply fine-tuning the model. 4) Furthermore, different masking strategies can impact the quality of synthetic clinical letters. Masking stopwords has a positive impact, while masking nouns or verbs has a negative effect. 5) For evaluation, BERTScore should be the primary quantitative evaluation metric, with other metrics serving as supplementary references. 6) Contextual information does not significantly impact the models' understanding, so the synthetic clinical letters have the potential to replace the original ones in downstream tasks.

Unlike previous research, which focuses more on restoring the original letters by training language models, this project provides a basic framework for generating diverse, de-identified clinical letters. It offers a direction for utilising the model to process real-world clinical letters, thereby helping to expand datasets in the clinical domain. Our codes and trained models are available at \url{https://github.com/HECTA-UoM/Synthetic4Health}


\end{abstract}


\section{Introduction}
\label{intro}
With the development of medical information systems, electronic clinical letters are increasingly used in communication between healthcare departments.  These clinical letters typically contain detailed information about patients' visits, including their symptoms, medical history, medications, etc \cite{rayner2020writing}. They also often include sensitive information, such as patients' names, phone numbers, and addresses \cite{tarur2021clinical,tucker2016protecting}.  As a result, these letters are difficult to share and nearly impossible to use widely in clinical education and research.

In 2018, 325 severe breaches of protected health information were reported by CynergisTek \cite{abouelmehdi2018big}. Among these, nearly 3,620,000 patients' records were at risk \cite{abouelmehdi2018big}. This is just the data from one year—similar privacy breaches are unfortunately common. The most severe hacking incident affected up to 16,612,985 patients \cite{abouelmehdi2018big}. Therefore, generating synthetic letters and applying de-identification techniques seem to be indispensable.

Additionally, due to privacy concerns and access controls, insufficient data is the major challenge of clinical education, medical research, and system development \cite{spasic2020clinical}. Some shared datasets offer de-identified annotated data. The MIMIC series is a typical example. These datasets are accessible through PhysioNet. MIMIC-IV \cite{johnson2023mimic,goldberger2000physiobank,johnson2024mimiciv}, the latest version, contains data from 364,627 patients' clinical information collected from 2008 to 2019 at a medical centre in Boston.  It records details about hospitalisations, demographics, and transfers. Numerous research are based on this shared dataset. Another public dataset series in the clinical domain is i2b2/n2c2 \cite{i2b2_n2c2}, They are accessible through the DBMI Data Portal. This series includes unstructured clinical notes such as process notes, radiology reports, and discharge summaries, and is published for clinical informatics sharing and NLP tasks challenges.

However, these sharing datasets are often limited to specific regions and institutions, making them not comprehensive. Consequently, models and medical research outcomes derived from these datasets cannot be widely applied \cite{humbert2022strategies}. Therefore, to address the lack of clinical datasets and reduce the workload for clinicians, it is essential to explore available technologies that can automatically generate de-identified clinical letters.

\begin{figure}[htbp]
  \centering
  \includegraphics[width=0.8\linewidth]{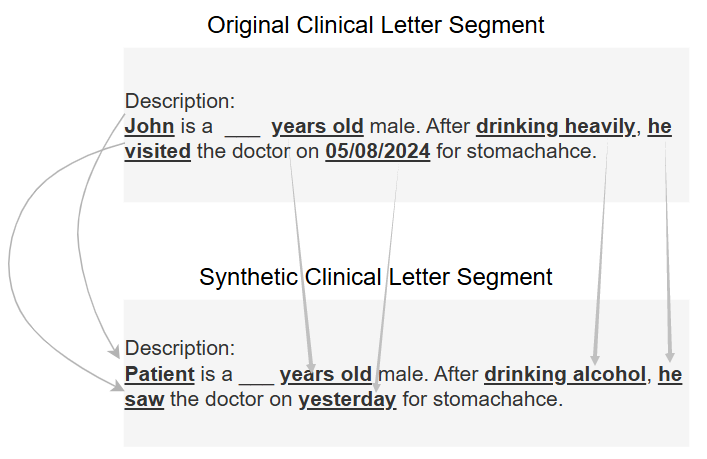}
  \caption{An Example of the Objective: sentence/segment-level generations}           

  \label{fig: An Example of the Objective}  
\end{figure}

Existing systems generate clinical letters primarily by integrating structured data, while there are not many studies on how to use Natural Language Generation (\textbf{NLG}) models for this task \cite{huske2003text,amin2020exploring,Tang2023DoesSD}. NLG attempts to combine clinical knowledge with general linguistic expressions and aims to generate clinical letters that are both readable and medically sound. However, NLG technology is not yet mature enough for widespread use in healthcare systems. Additionally, it faces numerous challenges, including medical accuracy, format normalisation, and de-identification \cite{huske2003text}. Therefore, this investigation focuses on how NLG technology can be used to generate reliable and anonymous clinical letters, which can benefit medical research, clinical education, and clinical decision-making.
The main aim of our work is to \textit{generate de-identified clinical letters} that can \textit{preserve clinical information} while \textit{differing from the original} letters. A brief example of our objective is shown in Figure \ref{fig: An Example of the Objective}.
Based on this objective, different generation models will be explored as a preliminary attempt. 
Then we select the best models and try various techniques to improve the quality of the synthetic letters. 
The synthetic letters are evaluated not only with quantitative and qualitative methods, but also in downstream tasks, i.e., Named Entity Recognition (NER). We hope this work will contribute to addressing the challenge of insufficient data in the clinical domain. 

In summary, this work is centered on the Research Question (RQ): "How can we generate reliable and diverse clinical letters without including sensitive information?" Specifically, it will answer the following related sub-questions (RQs):
\begin{enumerate}
\item How do different models perform in masking and generating clinical letters? 
\item How should the text be segmented in clinical letter generation?
\item How do different masking strategies affect the quality of synthetic letters?
\item How can we evaluate the quality of synthetic letters?
\end{enumerate}

To answer these questions, we explored various LLMs for masking and generating clinical letters, ultimately focusing on one that performed well. The overall \textbf{highlights} 
of this work are summarised as follows:

\begin{enumerate}
\item Mask and Generate clinical letters using different LLMs at the sentence level.
\item Explore methods to improve synthetic clinical letters' readability and clinical soundness.
\item Initially evaluate synthetic letters using both qualitative and quantitative methods.
\item Apply synthetic letters in downstream tasks.
\item Explore post-processing methods to further enhance the quality of de-identified letters.
\end{enumerate}

\section{Background and Literature Review}
\label{cha: Background and Literature review}

We first introduce general language models, followed by their applications, especially within the clinical domain. We then present the generative language models based on the Transformer architecture. These models serve as the technical foundation for most modern text generation tasks. Afterward, We review related works, discussing their relevance and how they are connected to ours.
Finally, all quantitative evaluation metrics used in this project are introduced.

    \subsection{Development of Language Models (LMs) }

The development of Language Models can be divided into three stages: Rule-Based Approach, Supervised Modelling, and Unsupervised Modelling \cite{boonstra2024artificial}.

    \subsubsection{Rule-Based Approach}
The Rule-Based Approach was first used in 1950s, which marks the beginning of NLP \cite{nadkarni2011natural}. This approach always uses a set of predefined rules, which were written and maintained manually by specialists \cite{van2018automated,satapathy2017sentiment}. Although it can generate standardised text without being fed with extensive input data \cite{van2018automated}, there are still numerous limitations. Initially, manually crafted rules are often ambiguous, and the dependencies between different rules increase maintenance costs \cite{nadkarni2011natural}.  Secondly, these stylised models cannot perform well in understanding realistic oral English and ungrammatical text such as clinical discharge records, although these texts are still readable for humans \cite{nadkarni2011natural}. Thirdly, they are not objective enough, as they are affected by the editors of the rule library. Additionally, they are not flexible enough to deal with special cases. Therefore, the Rule-Based Method is only suitable for analysing and generating highly standardised texts like prescriptions \cite{van2018automated}.
    
      \subsubsection{Supervised Language Models}
To address the limitations of the Rule-Based Approach, Supervised Learning has been applied to NLP. The invention of Statistical Machine Translation (SMT) in 1990 marked the rise of supervised NLP \cite{boonstra2024artificial}. It learns the correspondence rules between different languages by analysing the input of bilingual texts (parallel corpus) \cite{brown1990statistical}. Supervised NLP models are trained on annotated labels to learn rules automatically. The learned rules will be used in word prediction or text classification. 
Hidden Markov Model (HMM) and Conditional Random Field (CRF) are two typical applications of this stage \cite{sharma2023machine}. Both of them worked by tagging features of the input texts. HMM generates data by statistically analysing word frequencies \cite{eddy1996hidden,masuko1998text}. CRF, however, searches globally and calculates joint probabilities to get an optimal solution \cite{sutton2012introduction,bundschus2008extraction}.  Long Short-Term Memory (LSTM) is another typical example of supervised language modeling \cite{hochreiter1997long}.  
In the task of text generation, the input should be a set of labelled data or word vector sequences. By minimising the loss between the predicted word vector and the actual word vector, LSTM can capture the dependencies between words in long texts \cite{santhanam2020context,graves2012long}.

Although Supervised Language Models performs better than the Rule-Based Approach, domain experts still need to annotate the training dataset \cite{boonstra2024artificial}. In addition, data in some domains are difficult to collect due to privacy issues (such as \textbf{medical} and \textit{legal} domains). This became an ongoing challenge in applying the supervised language models to specific tasks.

      \subsubsection{Unsupervised Language Models}
To address the high cost and difficulty of obtaining labelled data, unsupervised Neural Networks are applied to the language modelling \cite{bengio2000neural}. The popularity of corpora such as Wikipedia and social media provides enough data for unsupervised models' training \cite{boonstra2024artificial}. Word embedding is a significant technique in this stage \cite{johnson2024detailed}. \textbf{Word2Vec} represents words using vectors with hundreds of dimensions. The context can be captured by training word vectors in the sliding window. By adjusting the hyperparameters to maximise the conditional probability of the target word, it can learn semantic information accurately \cite{mikolov2013efficient,mikolov2013distributed} (e.g. `Beijing'-`China'+`America' =\(>\) `Washington' \cite{ma2015using}).  After training, each word usually has a fixed word vector regardless of the context in which it appears (known as Static Word Embedding) \cite{santhanam2020context}. 

Unlike word2Vec, \textbf{BERT} and \textbf{GPT} use contextual word embedding. Their word vectors reflect the semantic information and are affected by the context \cite{camacho2018word}. BERT focuses on contextual understanding \cite{zhang2020semantics} (e.g. The bank is full of lush willows, where `bank' means the riverside, not the financial institution), while GPTs focus on text generation in a specific context \cite{liu2022makes,li2024pre} (e.g. Prompt: ``Do you know Big Ben?" Answer: ``Yes, I know Big Ben. It is the nickname for the Great Bell of the Clock located in London.") Although unsupervised language models have been able to train and understand text proficiently, they still face challenges in practical applications, such as difficulty handling ambiguity and high computing resource consumption. Therefore, language modelling still has a long way to go.

 \subsection{Language Models Applications in Clinical Domain}

Based on the modelling methods mentioned above, a variety of language models have been invented. They play an important role in scientific research and daily life, especially in the field of \textbf{healthcare}. In this section, I will discuss the \textit{clinical language model} applications in detail from two aspects: named entity recognition (\textbf{NER}) and natural language generation (\textbf{NLG}).
\subsubsection{Named Entity Recognition (NER)}

NER was originally designed for text analysis and recognition of named entities, such as dates, organisations, and proper nouns \cite{grishman1996message}. In the clinical domain, NER is used to identify \textit{clinical events} (e.g. symptoms, drugs, treatment plans, etc.) from unstructured documents with their \textit{qualifiers} (e.g. chronic, acute, mild), classify them, and extract the relationship \cite{bose2021survey,kundeti2016clinical}. Initially, NER relied on rule-based and machine-learning methods that required extensive manual feature engineering. In 2011, \newcite{collobert2011natural} used word embeddings and neural networks in NER. Since then, research in NER has shifted to automatic feature extraction.

\textbf{SpaCy} \footnote{\url{https://spacy.io/}} is an open-source NLP library for tasks like POS tagging and text classification. Additionally, it offers a range of pre-trained NER models. \textbf{ScispaCy} \footnote{\url{https://allenai.github.io/scispacy/}}, a fine-tuned extension of spaCy on medical science datasets, can recognise entities such as ``DISEASE”, ``CHEMICAL”, and ``CELL”, which are essential for medical research.
Although NER is useful in rapidly extracting clinical terms, there are still a lot of challenges, such as \textbf{non-standardisation} (extensive use of abbreviated words in clinical texts), \textbf{misspellings} (due to manual input by medical staff), and \textbf{ambiguity} (often influenced by context, e.g.whether 'back' refers to an adverb or an anatomical entity) \cite{bose2021survey}. Existing research mitigates these problems using \textit{entity linking} (mapping extracted clinical entities to medical repositories such as UMLS and SNOMED). More deep-learning models and text analysis tools are being developed to solve these issues.


\subsubsection{De-Identification}
The unprocessed clinical text has a risk of personal information leakage. Additionally, manual de-identification is not only prone to errors but also costly. Therefore, research on de-identification is indispensable for the secondary use of clinical data. Typically, de-identification is based on NER models to identify \textbf{Protected Health Information (PHI)}. Then PHI will be processed by different strategies (such as synonym replacement, removal, or masking) \cite{berg2021identification,berg2020impact}.

Similar to NER, early de-identification relied heavily on rule-based systems, machine learning, or hybrid models. Physionet DeID, the VHA Best-of-Breed (BoB), and MITRE’s MIST are three typical examples \cite{meystre2015identification}. However, these algorithms require extensive handcrafted feature engineering. With the development of unsupervised learning, recurrent neural networks (RNNs) and Transformers are widely used in de-identification tasks \cite{dernoncourt2017identification,kovavcevic2024identification}. 

\textbf{Philter}, a Protected Health Information filter \cite{norgeot2020protected},  is a pioneering system that combines rule-based approaches with state-of-the-art NLP models to identify and remove PHI. Although Philter outperforms many existing tools like Physionet and Scrubber, particularly in recall and F2 score, it still requires large amounts of annotated data for training \cite{norgeot2020protected}. Additionally, research has shown that while the impact of de-identification on downstream tasks is minimal, it cannot be completely ignored \cite{meystre2014text}. Therefore, performing de-identification without mistakenly removing semantic information is still a challenge in this field.

\subsubsection{Natural Language Generation (NLG)}
Both label-to-text and text-to-text generation are components of NLG \cite{gatt2018survey}. 
NLG consists of six primary sub-tasks, covering most of the NLG process. NLG architectures can generally be divided into three categories \cite{gatt2018survey}:
\begin{itemize}
    \item \textbf{Modular Architectures: }This architecture consists of three modules: the Text Planner (responsible for determining the content for generation), the Sentence Planner (which aggregates the synthetic text), and the Realiser (which generates grammatically correct sentences). These modules are closely related to the six sub-tasks, and each module operates independently.
    \item \textbf{Planning Perspectives: }This architecture considers NLG as a planning problem. It generates tokens dynamically based on the objectives, with potential dependencies between different steps.
    \item \textbf{Integrated or Global Approaches: }This is the dominant architecture for NLG, relying on statistical learning and deep learning. Common generative models, such as Transformers and conditional language models, are included in this architecture.
\end{itemize}

In the field of healthcare, NLG applications include document generation and question-answering. Document generation involves discharge letters, diagnostic reports for patients, decision-making suggestions for experts, and personalised patient profiles for administrators \cite{cawsey1997natural}. Some systems have already been implemented in practice. For instance, PIGLIT generates explanations of clinical terminology for diabetes patients \cite{hirst1997authoring}, and MAGIC can generate reports for Intensive Care Unit (ICU) patients \cite{mckeown1997language}. Question answering is another application of NLG. Tools like chatbots 
can provide patients with answers to basic healthcare questions \cite{locke2021natural}.  

Nowadays, NLG in the clinical field focuses on the development and training of transformer-based large language models (LLMs); examples of this work can be seen in \cite{amin2020exploring,luo2022biogpt}. These models perform well in specific domains such as semantic query \cite{kong2022transq} and electronic health records (\textbf{EHRs}) generation \cite{lee2018natural}. However, very few systems can reliably produce concise, readable, and clinically sound reports across multiple sub-domains \cite{cawsey1997natural}.

    \subsection{Generative Language Models}
    \label{sec: Generative Language Models}

    \subsubsection{Transformer and Attention Mechanism}
    \label{subsec: Transformer and Attention Mechanism}
Although Recurrent Neural Networks (RNNs) and Long Short-Term Memory (LSTM) are effective at semantic understanding, their recursive structure not only prevents parallel computation, but also makes them prone to gradient vanishing \cite{gillioz2020overview}. The introduction of the Transformer in 2017 addressed this issue by replacing the recurrent structure with a multi-head attention mechanism \cite{10.5555/3295222.3295349}. Since then, most deep learning models have been based on the Transformer. 
Transformer architecture is based on an encoder-decoder model \cite{10.5555/3295222.3295349}. To understand this, we first need to overview Auto-Regressive Models and the Multi-Head Attention Mechanism.

\textbf{Auto-Regressive Models}
Predictions for each Auto-Regressive Model token depend on the previous output. Therefore, it can only access the preceding tokens and operate iteratively. When the input sequence is \textit{X}, the Auto-Regressive Model aims to train parameters \(\mathit{\theta}\) to maximise the log-likelihood of the conditional probability \(\mathit{P}\) \cite{10.5555/3295222.3295349}.
\begin{equation}
L(X) = \sum_{i} \log P(x_i \mid x_{i-k}, \ldots, x_{i-1}; \Theta)
\end{equation}

\textbf{Multi-Head Attention Mechanism}
Attention mechanism was initially proposed by  \newcite{cho2014learning}. It can not only focus on the element being processed, but also capture the context dependence \cite{10.5555/3295222.3295349}. Multi-head attention consists of several single-head attention (Scaled Dot-Product Attention) layers \cite{10.5555/3295222.3295349}.
Each word in the input sequence is converted into a high-dimensional vector representing semantic information by word embedding.  These vectors pass linear transformation layers and get vectors for queries (\(\mathit{Q}\)), keys (\(\mathit{K}\)), and values (\(\mathit{V}\)). For each word, \(\mathit{Q}\), \(\mathit{K}\), and \(\mathit{V}\) are inputs to this single-head attention layer. 
The importance score of this word is calculated, and \(\mathit{V}\) corresponding to this word should be multiplied to get the output of this head (called Attention). Finally, outputs from all layers are concatenated to form a larger vector, which is the input to a feed-forward neural network (also the output of the multi-head attention layer) \cite{10.5555/3295222.3295349}.

\begin{equation}
\text{Attention}(Q, K, V) = \text{softmax}\left(\frac{QK^T}{\sqrt{d_k}}\right) V
\label{eq:attention}
\end{equation}

\textbf{Transformer and Pre-training Language Models (PLMs)}
\textbf{Transformer} consists of an encoder and a decoder. The Auto-Regressive Model is the basis of the decoder.  When the input sequence is \(X = (x_1, \ldots, x_N)\), output sequence is \(Y_M = (y_1, \ldots, y_M)\), the model can learn a latent feature representation  \(Z = (z_1, \ldots, z_N)\) from \(X\) to \(Y\). The generation of each new element \(Y_M\) relies on the generated sequence \(Y_{M-1} = (y_1, \ldots, y_{M-1})\) and feature representation \(Z\). Both the encoder and the decoder use the multi-head attention mechanism \cite{10.5555/3295222.3295349,gillioz2020overview}.

Many modern models are based entirely or partially on the Transformer. They compute general feature representations for the training set by unsupervised learning. This is the concept of Pre-training Language Models (PLMs). They can be fine-tuned to adapt to the specific tasks on particular datasets \cite{gillioz2020overview,li2024pre}. 

    \subsubsection{Encoder-Only Models}
Since the Transformer's encoder architecture can effectively capture the semantic features, some models only use this part for training. They are applied in text understanding tasks, such as text classification and NER. Bidirectional Encoder Representations from Transformers (BERT) \cite{devlin-etal-2019-bert} is a representative model among them. 

Unlike the Transformer decoder, which uses an Auto-Regressive model, BERT is trained based on the Masked Language Model (MLM) \cite{li2024pre}. It masks the word in the input sequence, and uses the bidirectional encoder to understand the context semantically, which will be used in predicting the masked word \cite{devlin-etal-2019-bert}. 
It has already been pre-trained on a 16GB corpus. 
To deploy it, we only need to replace the original fully connected layer with a new output layer, and then fine-tune the parameters on the dataset for specific tasks  \cite{devlin-etal-2019-bert}.  This approach consumes fewer computing resources and less time than training a model from scratch. In the clinical domain, Bio\_ClinicalBERT \cite{alsentzer2019publicly} and medicalai/ClinicalBERT \cite{wang2023optimized} are fine-tuned in the clinical dataset based on BERT architecture. Initially, due to the BERT's focus on semantic understanding, it was rarely used for text generation \cite{yang2019xlnet}. 

Robustly Optimized BERT Pretraining Approach (RoBERTa) \cite{zhuang-etal-2021-robustly} improved some key hyperparameters based on BERT. Instead of BERT's static mask, it uses a dynamic mask strategy, which helps it better adapt to multitasking. Additionally, it gained a stronger semantic understanding after training on five English datasets of 160GB. Unfortunately, compared to BERT, it requires significantly more computational resources and time \cite{acheampong2021transformer}.

To better handle long sequences, Longformer introduces a sparse attention mechanism to reduce computation \cite{beltagy2020longformer}. This allows each token to focus only on nearby tokens rather than the entire sequence. Unlike traditional models like BERT and RoBERTa, which can only process no more than 512 tokens, Longformer can handle up to 4096 tokens. It consistently achieves better performance than RoBERTa in downstream tasks involving long documents \cite{beltagy2020longformer}. The \textbf{Clinical-Longformer} model \cite{li2023comparative} was fine-tuned for the clinical domain.

Table \ref{Tab: Models and Datasets} summarises the \textbf{encoder-only models} used in our work and their corresponding fine-tuning datasets.

\begin{table}[htbp]
\centering
\begin{tabular}{|c|p{9cm}|}
\hline
\textbf{Model} & \textbf{Fine-tuned Dataset} \\ \hline
Bio\_Clinical BERT \cite{alsentzer2019publicly} & MIMIC-III \\ \hline
medicalai/ClinicalBERT \cite{wang2023optimized} & A large corpus of 1.2B words of diverse diseases \\ \hline
RoBERTa-base \cite{zhuang-etal-2021-robustly} & General Dataset (including BookCorpus, English Wikipedia, CC-News, OpenWebText, and Stories) \\ \hline
Clinical-Longformer \cite{li2023comparative} & MIMIC-III \\ \hline
\end{tabular}
\caption{Encoder-Only Models and Their Fine-tuned Datasets}
\label{Tab: Models and Datasets}
\end{table}
   
        
    \subsubsection{Decoder-Only Models}
In 2020, the performance of ChatGPT-3 \cite{10.5555/3495724.3495883} in question answering task caught researchers' attention to decoder-only architectures. 
As mentioned earlier, 
the Transformer decoder is an auto-regressive model. It can only refer to the synthesised words on the left side to generate the new word, without considering the context (which is called masked self-attention). This method made it more flexible in generating coherent text. Compared with BERT, the GPT series performed well in zero-sample and small-sample learning tasks by enlarging the size of the model. Even without fine-tuning, a simple prompt can help GPT generate a reasonable answer \cite{wu2024large}. 

Unlike GPT, which improves models' performance by increasing dataset size and the number of parameters without limitations, Meta AI published a series of \textbf{Llama} Models. These models aim to maximise the use of limited resources - in other words, by extending training, they reduce the overall demand on computing resources. The latest Llama3 model requires only 8B to 70B parameters \cite{llama3modelcard}, significantly fewer than GPT-3's 175 billion \cite{wu2024large}. Additionally, it outperforms GPT-3.5 Turbo in 5-shot learning \cite{web:contextai}.

    \subsubsection{Encoder-Decoder Models}

\textbf{T5} Family \cite{raffel2020exploring} is a classic example of the Encoder-Decoder Model. This architecture is particularly suitable for text generation tasks that require deep semantic understanding \cite{cai2022compare}. T5 transforms all kinds of NLP tasks into a text-to-text format \cite{tsirmpas2024neural}. Unlike BERT, which uses word-based masking and prediction, T5 processes text at the \textit{fragment} level using "span corruption" to understand semantics \cite{tsirmpas2024neural}.
For the fill-in-the-blank task, instead of replacing the specific words with \texttt{<mask>} like BERT, T5 replaces the text fragments with an ordered set of \texttt{<extra\_id\_n>} to reassemble the long sequence text. 
T5 needs to pre-process the input text according to the tasks' requirements. A directive prefix should be added as a prompt. 

Some language models fine-tuned with T5 on specific datasets, such as SciFive (fine-tuned in some science literature) \cite{Phan2021SciFiveAT} and ClinicalT5 (fine-tuned in clinical dataset MIMIC-III notes) \cite{lu-etal-2022-clinicalt5}, have shown excellent performance in their respective fields. The \textbf{T5 family models} used in this project and their corresponding fine-tuned datasets are summarised in Table \ref{Tab: T5 Family Models Comparison}.

%
\begin{table}[htbp]
\centering
\begin{tabular}{|p{5cm}|p{4cm}|p{4cm}|}\hline
\textbf{Model} & \textbf{Pre-trained Dataset} & \textbf{Weight Initialisation Method} \\ \hline
T5-base \cite{2020t5} & Colossal Clean Crawled Corpus (C4) & Randomly Initialised \\ \hline
Clinical-T5-Base \\ \cite{Lehman2023,doi:10.1161/01.CIR.101.23.e215}& MIMIC-\textsc{iii}, MIMIC-\textsc{iv} & Initialised from T5-Base \\ \hline
Clinical-T5-Sci \\ \cite{Lehman2023,doi:10.1161/01.CIR.101.23.e215}& PubMed Abstracts, PubMed Central & Initialised from SciFive \\ \hline
Clinical-T5-Scratch \\ \cite{Lehman2023,doi:10.1161/01.CIR.101.23.e215} & MIMIC-\textsc{iii}, MIMIC-\textsc{iv} & Randomly Initialised \\ \hline
\end{tabular}
\caption{The T5 Family Models used in our work}
\label{Tab: T5 Family Models Comparison}
\end{table}

    \subsubsection{Comparison and Limitations}
According to \newcite{cai2022compare}, the encoder-decoder architecture performs best with sufficient training data. However, challenges in data collection can negatively affect its performance. Despite these challenges, different architectures are well-suited to different tasks. For example, for tasks requiring semantic understanding, such as text summarisation, the encoder-decoder architecture is the most effective. In contrast, for tasks that involve minor word modifications, the encoder-only structure works better. However, the decoder-only structure is not suitable for tasks with insufficient training data and long text processing, but performs well in few-shot question-answering tasks \cite{amin-nejad-etal-2020-exploring,cai2022compare}.

Following these discussions, Transformer-based Pre-trained Language Models (PLMs) have demonstrated strong performance in NLP tasks, but many challenges still remain. 




   \subsection{Related Works on Clinical Text Generation}
    \label{sec: Related Works on Clinical Text Generation}
\subsubsection{LT3: Label to Text Generation}
\begin{figure}[htbp]
  \centering
  \includegraphics[width=0.8\linewidth]{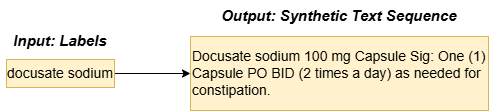}
  \caption{An Example of LT3}
  \label{fig: An Example of LT3}  
\end{figure}

LT3 \cite{belkadi2023generating} uses an encoder-decoder architecture to generate synthetic text from labels. As shown in Fig \ref{fig: An Example of LT3}, labels such as medications are the input of the encoder, which can generate corresponding feature representations. The decoder generates prescription sequences based on these features. The pre-trained BERT tokenizer is used to split the input sequence into sub-words. LT3 is trained from scratch. Instead of using traditional greedy decoding, which may miss the global optimum, the authors proposed Beam Search Decoding with Backtracking (\textbf{B2SD}). This approach broadens the search range through a backtracking mechanism, preserving possible candidates for the optimal solution. To reduce time complexity, they used a probability difference function to avoid searching for low-probability words. Additionally, the algorithm penalises repeated sub-sequences and employs a logarithmic heuristic to guide the exploration of generation paths. The authors test LT3 on the 2018-n2c2 dataset, and evaluate the results using both quantitative metrics and downstream tasks. It was demonstrated that this model outperforms T5 in label-to-text generation. \footnote{LT3 has shown significant improvements over existing label-to-text generation models. Unfortunately, when we tried applying B2SD to generate clinical letters, the results were somehow disappointing. This may be due to the length of clinical letters, B2SD consumes a lot of time on long text generation. Despite this, it still shows great potential in generating clinical data.
}


    \subsubsection{Seq2Seq Generation for Medical Dataset Augmentation}
\begin{figure}[htbp]
  \centering
  \includegraphics[width=0.8\linewidth]{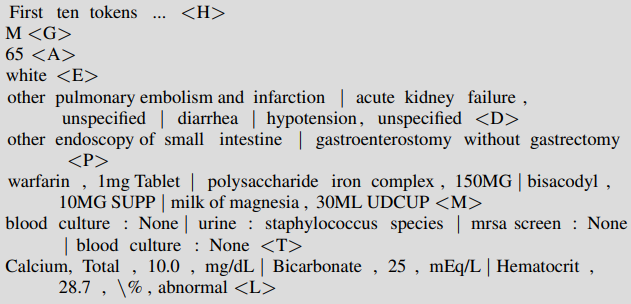}
  \caption{An Input Example of Conditional Text Generation}
  \label{fig: An Input Example of Conditional Text Generation}  
\end{figure}

\newcite{amin-nejad-etal-2020-exploring} compare the performance of the Vanilla Transformer and GPT-2 using the MIMIC-III dataset in seq2seq tasks. Specifically, they input a series of structured patient information as conditions, as shown in Fig \ref{fig: An Input Example of Conditional Text Generation}, to generate discharge summaries. They demonstrate that the augmented data outperforms the original data in downstream tasks (e.g. readmission prediction). Furthermore, they prove that Vanilla Transformer performs better with large samples, while GPT-2 excels in few-shot scenarios. However, GPT-2 is not suitable for augmenting long texts. Additionally, they used Bio\_ClinicalBERT for the downstream tasks, and discovered that Bio\_ClinicalBERT significantly outperformed the baseline model (BERT) in almost all experiments. It suggests that Bio\_ClinicalBERT can potentially replace BERT in the biomedical field. Interestingly, although the synthetic data have a low score on internal metrics (such as ROUGE and BLEU), the performance on downstream tasks is notably enhanced. This may be because augmenting text can effectively introduce noise into the original text, improving the model's generalisation to unseen data.

According to their findings, decoder-only models like GPT-2 are not suitable for processing long text. Bio\_ClinicalBERT is particularly effective for tasks in the clinical area, and Clinical Transformer is promising in augmenting medical data. This provides more possibilities for my task of generating synthetic clinical letters.

\subsubsection{Discharge Summary Generation Using Clinical Guidelines and Human Evaluation Framework}
Unlike the traditional supervised learning of fine-tuning language models (which requires a large amount of annotated data), \newcite{ellershaw2024automated} generated 53 discharge summaries using only a one-shot example and a clinical guideline. Their research consists of two aspects: generating discharge summaries and a manual evaluation framework.

\begin{figure}[htbp]
  \centering
  \includegraphics[width=0.8\linewidth]{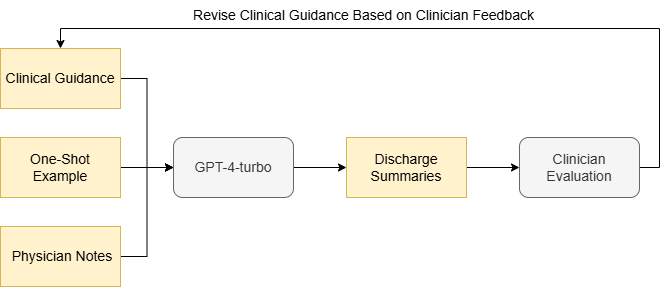}
  \caption{Workflow of Discharge Summary Generation Using Clinical Guidelines}
  \label{fig: Workflow of Discharge Summary Generation Using Clinical Guidelines}  
\end{figure}

As shown in Fig \ref{fig: Workflow of Discharge Summary Generation Using Clinical Guidelines}, the authors used clinical notes from MIMIC-III as input, and incorporated a one-shot summary along with clinical guidance as prompts to generate discharge summaries by GPT-4-turbo. Initially, five sample synthetic summaries were evaluated by a clinician. Based on the feedback, the clinical guidance was revised to adapt to the generation task. Through iterative optimisation, the revised guidance, combined with the original one-shot sample, became the new prompt. Then the authors generated 53 discharge summaries using this method and invited 11 clinicians to do a final manual quantitative evaluation.
Clinicians were invited to evaluate the error rate at the section level (e.g., Diagnoses, Social Context, etc). It includes four dimensions:

\begin{itemize}
    \item Minor omissions;
    \item Severe omissions;
    \item Unnecessary text;
    \item Incorrect additional text.
\end{itemize}

Each discharge summary was evaluated by at least two clinicians, and the authors calculated agreement scores to evaluate the subjectivity during the human evaluation stage. Unfortunately, the inter-rater agreement was only 59.72\%, raising concerns that the revised prompts based on such feedback might result in subjective synthetic summaries. 
Although this study partially addresses the issue of insufficient training data, and provides reliable human quantitative evaluation methods, it is still not well-suited for our investigation. Specifically, 
It consumes significant time and manpower.
Therefore, there is still a long way to go before this technique can be used for large-scale text generation tasks.

\subsubsection{Comparison of Masked and Causal Language Modelling for Text Generation}

\begin{figure}[htbp]
  \centering
  \includegraphics[width=0.8\linewidth]{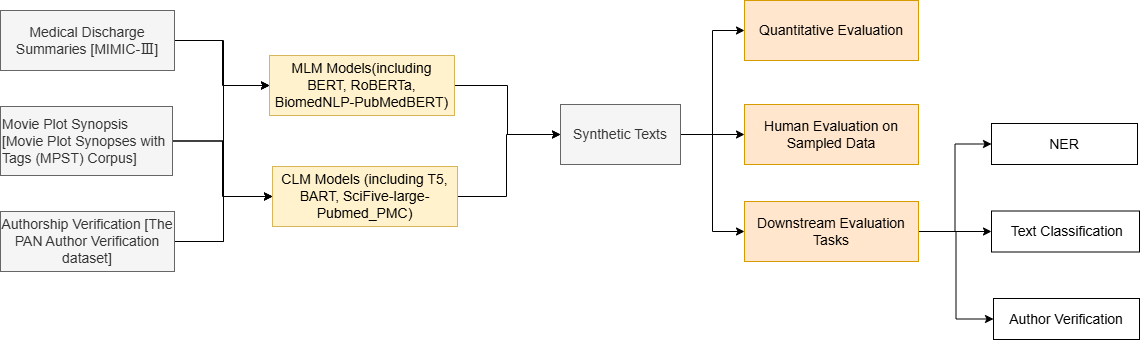}
  \caption{Workflow of MLM and CLM Comparison in Text Generation}
  \label{fig: Workflow of MLM and CLM Comparison}  
\end{figure}

\newcite{DBLP:journals/corr/abs-2405-12630} compared masked language modelling (MLM, including BERT, RoBERTa, BiomedNLP-PubMedBERT) and causal language modelling (CLM, including T5, BART, SciFive-large-Pubmed\_PMC) across various datasets in masking and text generation tasks. They used qualitative and quantitative evaluations, as well as downstream tasks, to assess the quality of the synthetic texts. Their workflow is shown in Fig \ref{fig: Workflow of MLM and CLM Comparison}. Based on these evaluations, the study yielded the following results:
\begin{itemize}
    \item MLM models are better suited for text masking and generation tasks compared to CLM.
    \item Introducing domain-specific knowledge does not significantly improve the model's performance.
    \item Downstream tasks can adapt to the introduced noise. Although some synthetic texts might not achieve high quantitative evaluation scores, they can still perform well in downstream tasks. This matches the findings from Amin-Nejad \cite{amin-nejad-etal-2020-exploring}.
    \item A lower random masking ratio can generate higher-quality synthetic texts.
\end{itemize}
These very recent findings provide insightful inspiration to our investigation. Our work will build on their research, expanding on masking strategies and focusing on the clinical domain.


\section{Methodologies and Experimental Design} 

\begin{figure}[htbp]
  \centering
  \includegraphics[width=0.99\linewidth]{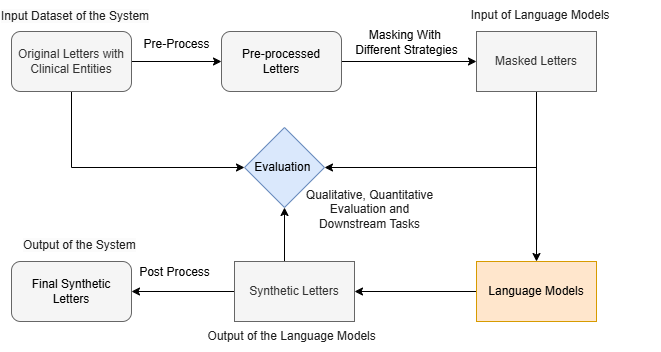}
  \caption{Overall Investigation Workflow for \textsc{Synthetic4Health}}           
  \label{fig: Overall Workflow}  
\end{figure}

Due to the sensitivity of clinical information, many clinical datasets are not accessible. As mentioned in Section \ref{cha: Background and Literature review}, numerous studies use NLG techniques to generate clinical letters, and evaluate the feasibility of replacing the original raw clinical letters with synthetic letters. Most existing research involves fine-tuning PLMs or training Transformer-based models from scratch on their datasets through supervising learning.
These studies explore different ways to learn the mapping from original raw text to synthetic text and work on generating synthetic data that are similar (or even identical) to the original ones. Our work, however, aims to find a method that can generate clinical letters that can \textit{keep the original clinical story, while not exactly being the same as the original letters}. To achieve this objective, we employed various models and masking strategies to generate clinical letters. The experiment will follow these steps:

\begin{enumerate}
\item \textbf{Data Collecting and Pre-processing}: Access clinical letter examples \cite{goldberger2000physiobank,johnson2024mimiciv,johnson2023mimic}. Segment the text at sentence level. Extract entities and the letters' templates to represent the clinical story and maintain clinical soundness.
\item \textbf{Randomly mask} the context. Generate clinical letters by predicting masked tokens using different LLMs. 
\item \textbf{Evaluate} synthetic letters generated by different language models. Select one well-performed model - Bio\_ClinicalBERT, and work on it.
\item Explore different \textbf{masking strategies} to retain clinical stories and diversity while removing private information. After generating clinical letters using these strategies, evaluate their quality.
\item Explore \textbf{post-processing} methods, in order to further enhance the readability of synthetic letters.
\item Compare the performance of synthetic and original letters in a \textbf{downstream NER} task, to evaluate the usability of these synthetic letters.
\end{enumerate}
An overall investigation workflow is shown in Fig \ref{fig: Overall Workflow}.

    

    \subsection{Data Set}
    \label{sec: Experimental Setup}

Based on the objective of this project, we need a dataset that includes both clinical \textbf{notes} and some clinical \textbf{entities}. The dataset we used is from the SNOMED CT Entity Linking Challenge \cite{goldberger2000physiobank,johnson2024mimiciv,johnson2023mimic}. It includes 204 clinical letters and 51,574 manually annotated clinical entities. 

\textbf{Clinical Letters} 
The clinical letters are from a subset of discharge summaries in MIMIC-IV-Note \cite{goldberger2000physiobank,Johnson2023}. It uses clinical notes obtained from a healthcare system in the United States. These notes were de-identified by a hybrid method of the Rule-based Approach and Neural Networks. To avoid releasing sensitive data, the organisation also did a manual review of protected health information (PHI). In these letters, all PHI was replaced with three underscores ‘\textunderscore\textunderscore\textunderscore’. The letters record the patient's hospitalisation information (including the reason for visiting, consultation process, allergy history, discharge instructions, etc.). They are saved in a comma-separated value (CSV) format file `mimic-iv\_notes\_training\_set.csv'. Each row of data represents an individual clinical letter. It consists of two columns, where the "note\textunderscore id" column is a unique identifier for each patient's clinical letter, and the ‘text’ column contains the contents of the clinical letter. Since most language models have a limitation on the number of tokens to process \cite{sun2021revisiting}, we tokenized the clinical letters into words using the `NLTK' library and found that all clinical letters contained thousands of tokens. Therefore, it is necessary to split each clinical letter into multiple chunks to process them. These separated chunks should be merged in the end to generate the whole letter.

\textbf{Annotated Clinical Entities} 
The entities are manually annotated based on SNOMED CT. A total of 51,574 annotations cover 5,336 clinical concepts. They are saved in another CSV document which includes four columns: `note\_id', `start', `end', and `concept\_id'. The `note\_id' column corresponds to the `note\_id' in `mimic-iv\_notes\_training\_set.csv' file. The `start' and `end' columns indicate the position of annotated entities. The `concept\_id' can be used for entity linking with SNOMED CT. For example, for the `note\_id'`10807423-DS-19', , the annotated entity `No Known Allergies' has a corresponding `concept\_id': `609328004'. This can be linked to SNOMED CT under the concept of `Allergic disposition' \cite{SNOMEDCT}.

\begin{figure}[htbp]
  \centering
  \includegraphics[width=0.99\linewidth]{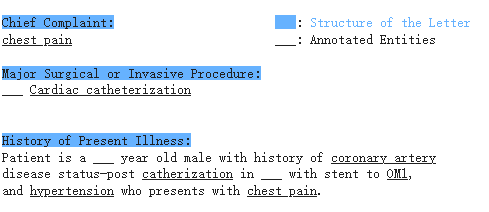}
  \caption{Text Excerpt from the Original Letter \cite{goldberger2000physiobank,johnson2024mimiciv,johnson2023mimic} (`note\_id': '17656866-DS-6')}           
  \label{fig: Text Excerpt from the Original Letter}  
\end{figure}

An example of text excerpted from the original letter is shown in Figure \ref{fig: Text Excerpt from the Original Letter}. It contains the document structure and some free text. According to the dataset, document structure often corresponds to capital letters and colons `:’. Our primary goal is to mask the context that is neither part of the document structure nor annotated entities, and then generate a new letter, as both \textbf{structure} and clinical \textbf{entities} are essential for understanding clinical information \cite{meystre2014text}.


 
    \subsection{Software and Environment}
\hspace{1em}All codes and experiments in this project are run in the integrated development environment (IDE) `Google Colab'. The built-in T4 GPU is used to accelerate the inference process. The primary tools used in the project include:
\begin{itemize}
    \item \textbf{Programming Language and Environment: }Python 3.8 serves as the main programming language.
    \item \textbf{Deep Learning Framework: }PyTorch is the core framework used for loading and applying pre-trained language models (PLMs).
    \item \textbf{Natural Language Processing Libraries: }This includes Hugging Face's Transformers, NLTK, BERTScore, etc. They are popular tools for text processing and evaluation in the NLP domain.
    \item \textbf{Auxiliary Tools: }Libraries such as Pandas and Math can support data management, mathematical operations, and other routine tasks.
\end{itemize}

    \subsection{Pre-Processing}
    \label{sec: Pre-Processing}

\begin{figure}[htbp]
  \centering
  \includegraphics[height=0.25\textheight]{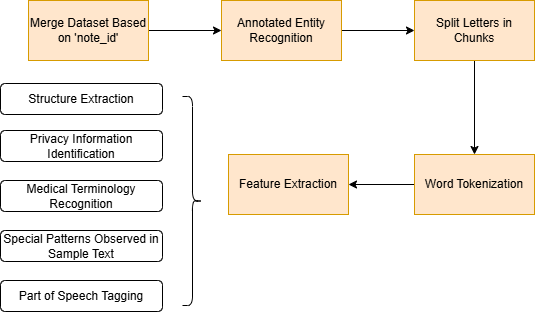}
  \caption{Pre-Processing Pipeline}           
  \label{fig: Pre-Processing Pipeline}  
\end{figure}

The collected dataset involves different files and is entirely raw data. It is necessary to pre-process them before using them in generation tasks. The pre-processing of this system contains five steps: `Merge dataset based on `note\_id'', `Annotated Entity Recognition', `Split Letters in Chunks', `Word Tokenization' and `Feature Extraction'. The pre-processing pipeline is shown in Fig \ref{fig: Pre-Processing Pipeline}.

    \subsubsection{Merging Dataset and Annotated Entity Recognition}
\begin{figure}[htbp]
  \centering
  \includegraphics[height=0.25\textheight]{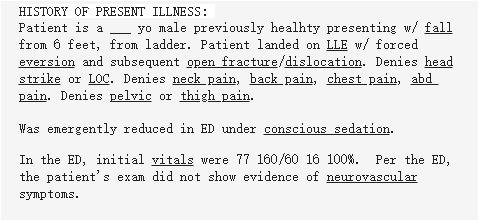}
  \caption{Sample Text from Original Letters \cite{goldberger2000physiobank,johnson2024mimiciv,johnson2023mimic} (`note\_id': '10807423-DS-19')}           
  \label{fig: Sample Text from Original Letters}  
\end{figure}

Initially, we merged the clinical letters file and annotations file into a new DataFrame. 
The method is detailed in Appendix \ref{Appendix: Method for Dataset Merging in Preprocessing}. 
After this, we extracted manually annotated entities based on their index. An excerpt from an original letter is shown in Fig \ref{fig: Sample Text from Original Letters}, and the manually annotated entities are listed in Table \ref{Tab: Extracted Annotated Entities}.

\begin{table}[htbp]
\centering
\begin{tabular}{|c|c|c|c|}
\hline
\textbf{Entity} & \textbf{Start} & \textbf{End} & \textbf{Concept ID} \\ \hline
fall & 571 & 575 & 161898004 \\ \hline
LLE & 621 & 624 & 32153003 \\ \hline
eversion & 636 & 644 & 4196002 \\ \hline
open fracture & 660 & 673 & 397181002 \\ \hline
dislocation & 674 & 685 & 87642003 \\ \hline
head strike & 694 & 706 & 82271004 \\ \hline
LOC & 710 & 713 & 419045004 \\ \hline
neck pain & 722 & 731 & 81680005 \\ \hline
back pain & 733 & 742 & 161891005 \\ \hline
chest pain & 744 & 754 & 29857009 \\ \hline
abd pain & 756 & 765 & 21522001 \\ \hline
pelvic & 774 & 780 & 30473006 \\ \hline
thigh pain & 784 & 794 & 78514002 \\ \hline
conscious sedation & 833 & 851 & 314271007 \\ \hline
vitals & 874 & 880 & 118227000 \\ \hline
neurovascular symptoms & 963 & 986 & 308921004 \\ \hline
\end{tabular}
\caption{Extracted Entities and Their Details}
\label{Tab: Extracted Annotated Entities}
\end{table}

    \subsubsection{Splitting Letters into Variable-Length Chunks}
Typically, PLMs such as BERT, RoBERTa, and T5 have a limit on the number of input tokens, usually capped at 512 \cite{zeng2022not}. When dealing with text that exceeds this limit, common approaches include discarding the excess tokens or splitting the text into fixed-length chunks of 512 tokens. In addition, some studies evaluate the tokens' importance to decide which parts should be discarded \cite{hou2022token}.

In this work, each clinical letter (`note\_id') contains thousands of tokens, as mentioned in Section \ref{subsec: Dataset}, to preserve as much critical clinical information as possible, we avoided simply discarding tokens. 
Instead, we adopted a splitting strategy based on \textbf{semantics}. Each block is not a fixed length. Rather, they are complete \textbf{paragraphs} that are as close as possible to the token limit. This approach aims to help the model better capture the meaning and structure of clinical letters, thereby improving its ability to retain essential clinical information while efficiently processing the text. In fact, we initially generated letters at the sentence level. However, it was found that processing at the sentence level is not only time-consuming, but also fails to provide the model with enough information for inference and prediction. This is why the letters are processed in chunks rather than in sentences.

\begin{figure}[htbp]
  \centering
  \includegraphics[width=\textwidth]{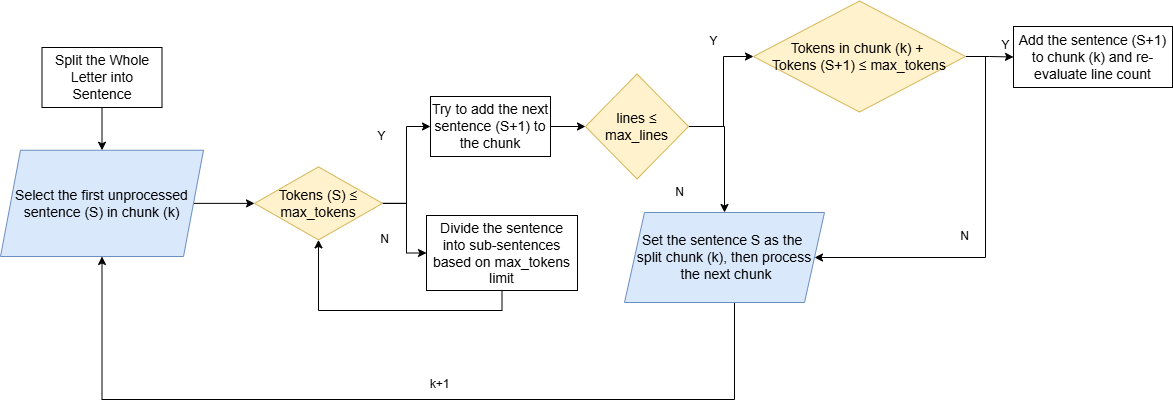}
  \caption{Text Chunking Workflow}           
  \label{fig: Text Chunking Workflow}  
\end{figure}

As shown in Fig \ref{fig: Text Chunking Workflow}, each raw letter is split into sentences first. We used the pre-trained models provided by the ‘NLTK’ library, which combines statistical and machine-learning approaches to identify sentence boundaries. Each clinical letter is treated as a separate processing unit, with the first sentence automatically assigned to the first text block (chunk). To control the length of each chunk, we set a \textbf{maximum line count} parameter (max\_lines). If the first sentence already meets the value of `max\_lines', the chunk will only contain this one sentence. Otherwise, subsequent sentences will be added to the chunk until the line count up to the max\_lines.

\begin{figure}[htbp]
  \centering
  \includegraphics[width=0.7\textwidth]{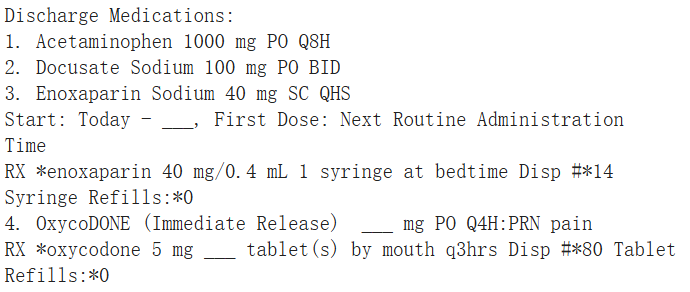}
  \caption{Sentence Fragment Exceeding Token Limit \cite{goldberger2000physiobank,johnson2024mimiciv,johnson2023mimic} (`note\_id': '10807423-DS-19')}            
  \label{fig: Excerpt of a Sentence Beyond Token Limitation}  
\end{figure}

Extra care is needed when handling text with specific formats, such as medication dosage descriptions, as shown in Fig \ref{fig: Excerpt of a Sentence Beyond Token Limitation}. Because there is no clear sentence boundary, these sentences may exceed the tokens limitation. To address this, we first check whether the sentence being processed exceeds the token limit (max\_tokens). If it does not, the sentence will be added to the current chunk. Otherwise, the sentence should be split into smaller chunks, each no longer than `max\_tokens'. This operation helps \textbf{balance processing efficiency} while maintaining semantic integrity. In the example shown in \ref{fig: Excerpt of a Sentence Beyond Token Limitation}, although using line breaks to split the text seems to be more flexible, considering time complexity and the requirement to index the annotated entities, this method was not chosen.

    \subsubsection{Word Tokenisation}
To prepare the text for model processing, we split each chunk of text into smaller units: tokens. The tokenisation methods can be categorised into two types: one for feature extraction and the others for masking and generation.

For the tokenization aimed at feature extraction, we used the `word\_tokenize' method from the `NLTK' library. It is helpful to preserve the original features of the words, which is especially important for retaining clinical entities. For instance, in the sentence ``Patient is a \_\_\_ yo male previously healthy presenting w/ fall from 6 feet, from ladder." Word boundaries such as spaces can be automatically detected for tokenization. The results of different tokenization methods are shown in the Table \ref{Tab: Comparison of Tokenization Methods for Different LMs}.

As for the tokenization used for masking and generating, we retained the original models' tokenization methods. The specific tokenization approach varies by model, as shown in Table \ref{Tab: Comparison of Tokenization Methods for Different LMs}. For example, BERT family models use Word-Piece tokenization, which initially splits text by spaces and then further divides the words into sub-words \cite{zhuang-etal-2021-robustly}. This approach is particularly effective for handling words that are not in the pre-training vocabulary and is especially useful for predicting masked words. For complex clinical terms, however, these models rely heavily on a predefined dictionary, which can result in unsatisfactory tokenization and hinder the model’s understanding. For instance, the word `COVID-19’ is tokenized by BERT into [`co', `\#\#vid', `-', `19']. In contrast, the T5 family models use Sentence-Piece tokenization. It does not rely on space to split the text. Instead, this method tokenizes directly from the raw text, making it better suited for handling abbreviations and non-standard characters (e.g. `COVID-19'),  which are common in clinical letters.

It is important to note that although all BERT family models use Word-Piece tokenization, the results can still differ. This is because different models use different vocabularies during pre-training, leading to variations in tokenization granularity. The tokenization methods for each model are detailed in Table \ref{Tab: Comparison of Tokenization Methods for Different LMs}. Each tokenization approach has its own advantages and disadvantages for processing clinical letters. Therefore, exploring how these models impact the clinical letter generation is also a requirement of my project.

\begin{table}[htbp]
\centering
\begin{tabular}{|p{4cm}|p{3cm}|p{8cm}|}
\hline
\textbf{Operation / Model} & \textbf{Tokenization Method} & \textbf{Tokenized Output} \\ \hline
Feature Extraction & Word Tokenization & 
[`Patient', `is', `a', `\_\_\_', `yo', `male', `previously', `healthy', `presenting', `w\/', `fall', `from', `6', `feet', `,', `from', `ladder', `.]
\\ \hline
medicalai / ClinicalBERT & Subword-Enhanced Word-Piece & [`patient', `is', `a', `\_', `\_', `\_', `yo', `male', `previously', `healthy', `presenting', `w', `/', `fall', `from', `6', `feet', `,', `from', `la', `\#\#dder', `.]
\\ \hline
BERT-base, Bio\_ClinicalBERT & Standard Word-Piece & 
[\textquotesingle patient\textquotesingle, \textquotesingle is\textquotesingle, \textquotesingle a\textquotesingle, \textquotesingle\_\textquotesingle, \textquotesingle\_\textquotesingle, \textquotesingle\_\textquotesingle, \textquotesingle yo\textquotesingle, \textquotesingle male\textquotesingle, \textquotesingle previously\textquotesingle, \textquotesingle healthy\textquotesingle, \textquotesingle presenting\textquotesingle, \textquotesingle w\textquotesingle, \textquotesingle/\textquotesingle, \textquotesingle fall\textquotesingle, \textquotesingle from\textquotesingle, \textquotesingle 6\textquotesingle, \textquotesingle feet\textquotesingle, \textquotesingle,\textquotesingle, \textquotesingle from\textquotesingle, \textquotesingle ladder\textquotesingle, \textquotesingle.\textquotesingle] 
\\ \hline
Clinical-Longformer, RoBERTa & Detailed WordPiece & [\textquotesingle Pat\textquotesingle, \textquotesingle ient\textquotesingle, \textquotesingle Ġis\textquotesingle, \textquotesingle Ġa\textquotesingle, \textquotesingle Ġ\_\_\_\textquotesingle, \textquotesingle Ġyo\textquotesingle, \textquotesingle Ġmale\textquotesingle, \textquotesingle Ġpreviously\textquotesingle, \textquotesingle Ġhealthy\textquotesingle, \textquotesingle Ġpresenting\textquotesingle, \textquotesingle Ġw\textquotesingle, \textquotesingle/\textquotesingle, \textquotesingle Ġfall\textquotesingle, \textquotesingle Ġfrom\textquotesingle, \textquotesingle Ġ6\textquotesingle, \textquotesingle Ġfeet\textquotesingle, \textquotesingle,\textquotesingle, \textquotesingle Ġfrom\textquotesingle, \textquotesingle Ġladder\textquotesingle, \textquotesingle.\textquotesingle] \\ \hline
T5 Family (T5 base, SCI T5, Clinical T5) & Sentence-Piece & 
[`\_\_Patient', `\_\_is', `\_\_', `a', `\_\_', `\_', `\_', `\_', `\_\_', `y', `o', `\_\_male', `\_\_previously', `\_\_healthy', `\_\_', `presenting', `\_\_', `w', `/', `\_\_fall', `\_\_from', `\_\_6', `\_\_feet', `,', `\_\_from', `\_\_ladder', `.’]
\\ \hline
\end{tabular}
\caption{Comparison of Tokenization Methods for Different LMs on Sentence \\ "Patient is a \_\_\_ yo male previously healthy presenting w/ fall from 6 feet, from ladder."}
\label{Tab: Comparison of Tokenization Methods for Different LMs}
\end{table}

    \subsubsection{Feature Extraction}
\begin{figure}[htbp]
  \centering
  \includegraphics[width=0.7\textwidth]{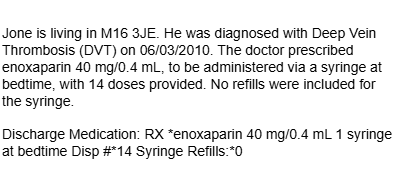}
  \caption{Example Sentence for Feature Extraction (See Table \ref{Tab: Feature Extraction} for Extracted Features)}           
  \label{fig: Feature extraction example}  
\end{figure}

Since we aim to generate de-identified clinical letters that can preserve clinical narratives during masking and generation, it is necessary to extract certain features beforehand.
We extracted the following features, with an example provided in Figure \ref{fig: Feature extraction example} and Table \ref{Tab: Feature Extraction}.

\begin{itemize}
    \item \textbf{Document Structure: }This feature is identified by a rule-based approach. As mentioned in Subsection \ref{subsec: Dataset}, structural elements (or templates) often correspond to the use of colons ‘:’. They should not be masked to preserve the clinical context.
    \item \textbf{Privacy Information Identification: }In this part, I used a hybrid approach. To identify sensitive information such as `Name', `Date', and `Location (LOC)', I employed a NER toolkit from Stanza \cite{qi2020stanza}. To handle privacy information like phone numbers, postal codes, and e-mail addresses, I implemented a rule-based approach. Specifically, I devised several regular expressions to match the common formats of these data types. These identified privacy information should be masked.
    \item \textbf{Medical Terminology Recognition: }A NER toolkit pre-trained on the dataset i2b2 is used here \cite{zhang2021biomedical}. It can identify terms like `Test', `Treatment', and `Problem' in free text. Although our dataset has already been manually annotated, these identified terms can serve as a supplement to the pre-annotated terms.
    \item \textbf{Special Patterns Observed in Sample Text: }Some specific patterns, like medication dosages (e.g. enoxaparin 40 mg/0.4 mL) or special notations (e.g. `b.i.d.'), may carry significant meaning. I retained these terms unless they were identified as private information to preserve the clinical background of the raw letters.
    \item \textbf{Part of Speech (POS) Tagging: }Different parts of speech (POS) play distinct roles in interpreting clinical texts. I aim to explore how these POS influence the model’s understanding of clinical text. To achieve this, I used a toolkit \cite{zhang2021biomedical} trained on the MIMIC-III \cite{johnson2016mimic} dataset for POS tagging. 
    It performs better than SpaCy \footnote{\url{https://spacy.io/}} and NLTK in handling clinical letters.
\end{itemize}


\begin{table}[htbp]
\centering
\begin{tabular}{|p{6cm}|p{8cm}|}
\hline
\textbf{Feature Extraction Operation} & \textbf{Extracted Features} \\ \hline

Structure Extraction & Discharge Medication: \\ \hline

Privacy Information Identification & 
Jone (PERSON) \newline
06/03/2010 (DATE) \newline
Postal Code: M16 3JE \\ \hline

Medical Terminology Recognition & 
Deep Vein Thrombosis (PROBLEM) \newline
DVT (PROBLEM) \newline
enoxaparin (TREATMENT) \newline
the syringe (TREATMENT) \newline
RX \*enoxaparin (TREATMENT) \\ \hline

Special Patterns Observed in Sample Text & 
`40 mg/0.4 mL' (122, 134) \newline
`40 mg/0.4 mL ' (284, 297) \newline
`\#\*14' (323, 327) \newline
`\*0' (344, 346) \\ \hline

POS Tagging & 
`Jone', `PROPN' \newline
`is', `AUX' \newline
`living', `VERB' \newline
`in', `ADP' \newline 
\textit{(Partial List)} \\ \hline

\end{tabular}
\caption{Example: Summary of Feature Extraction Operations and Extracted Features}
\label{Tab: Feature Extraction}
\end{table}

  \subsection{Clinical Letters Generation}
    \label{sec: Clinical Letters Generation}
 We discuss the models and masking strategies that are used in generating synthetic clinical letters. It is important to clarify that our key objective is to generate letters that differ from the original ones, rather than being exact copies, as the same statement may indirectly reveal the patients' privacy. Although fine-tuning the model can always improve precision and enhance the model's semantic comprehension ability, it tends to produce letters that are too closely aligned with the originals. This also causes the fine-tuned model to rely too heavily on the original dataset, compromising its ability to generalise. Therefore, simply fine-tuning the model is not ideal if the PLMs can already generate the readable text. Instead, we should concentrate on how to \textit{protect clinical terms and patient narratives as well as avoid privacy breaches}.

As discussed in Section \ref{sec: Generative Language Models} and Section \ref{sec: Related Works on Clinical Text Generation}, decoder-only models struggle with processing long texts that require contextual understanding \cite{amin-nejad-etal-2020-exploring}. Additionally, deploying them requires substantial computing resources and time. Therefore, we explored various pre-trained language models (PLMs), including both encoder-only and encoder-decoder models in this project. After evaluating their ability to generate synthetic letters from our dataset, we focused on \textbf{Bio\_ClinicalBERT} - a well-performed model in our task, to experiment with different masking strategies. Additionally, from the discussion in  Section \ref{sec: Pre-Processing}, we need to split the text into various-length-chunks. So the appropriate \textit{length of these chunks} is also experimented with Bio\_ClinicalBERT.

    \subsubsection{Encoder-Only Models  with Random Masking}
 As mentioned earlier, the primary method for this project involves masking and generation. We focused extensively on encoder-only models because of their advantage in bi-directional semantic comprehension. These encoder-only models, including BERT, RoBERTa, and Longformer—detailed in Section \ref{sec: Generative Language Models}—were compared for their performance. Given the clinical focus of this task, we particularly explored model variants that were fine-tuned on clinical or biological datasets. However, as no clinically fine-tuned RoBERTa \cite{zhuang-etal-2021-robustly} variant was available, the \textbf{RoBERTa-base} was used for comparisons. Specifically, the encoder-only models we explored include Bio\_ClinicalBERT \cite{alsentzer2019publicly}, medicalai/ClinicalBERT \cite{wang2023optimized}, RoBERTa-base  \cite{zhuang-etal-2021-robustly}, and Clinical-Longformer  \cite{li2023comparative}.

We used the standard procedure for Masked Language Modelling (\textbf{MLM}). First, the tokens that need to be masked are selected. They are then corrupted, resulting in masked text - that includes both masked and unmasked tokens. Next, the model predicts the masked tokens and replaces them with the ones that have the highest probabilities. 

    \subsubsection{Encoder-Decoder Models  with Random Masking}
 Although encoder-decoder models are not typically used for masked language modelling, they are well-suited for text generation. The architecture of T5, in particular, is designed to maintain the coherence of the text \cite{raffel2020exploring}. Therefore, we included the T5 family models in comparisons.  

\begin{figure}[htbp]
  \centering
  \includegraphics[width=0.7\textwidth]{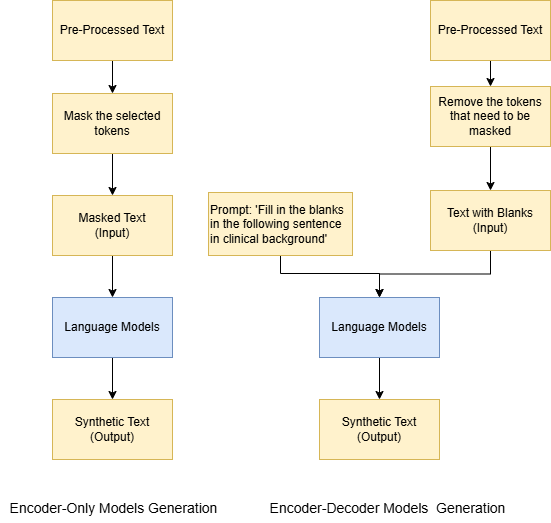}
  \caption{Comparison of Encoder-Only and Encoder-Decoder Model Architectures}           
  \label{fig: Comparison of Encoder-Only and Encoder-Decoder Model Architectures}  
\end{figure}

The process of generating synthetic letters with encoder-decoder models is very similar to that with encoder-only models. The difference is that, unlike the BERT family, which automatically masks tokens and replaces them with ’\texttt{<mask>}‘, the T5 family models \textit{do not have any built-in masking function}. As a result, we identified the words that needed to be masked by \textbf{index} and removed them, which are represented as `extra\_id\_x' in the T5 family models. The text, with these words removed, was then used for the generation, which we refer to as `\textbf{text with blanks}'. To maintain consistency in the format, we later replaced `extra\_id\_x' with `\texttt{<mask>}' when displaying the masked text.  Additionally, the T5 family models \textbf{require a prompt} as part of the input. For this task, the complete input was structured as `Fill in the blanks in the following sentence in the clinical background' + `text with blanks'. In this project, we used T5-base \cite{2020t5}, Clinical-T5-Base \cite{Lehman2023,doi:10.1161/01.CIR.101.23.e215}, Clinical-T5-Sci \cite{Lehman2023,doi:10.1161/01.CIR.101.23.e215}, and Clinical-T5-Scratch \cite{Lehman2023,doi:10.1161/01.CIR.101.23.e215} for comparison. The comparison of encoder-only and encoder-decoder model architectures is shown in Fig \ref{fig: Comparison of Encoder-Only and Encoder-Decoder Model Architectures}.

    \subsubsection{Different Masking Strategies with Bio\_ClinicalBERT}
    \label{subsec: Different Masking Strategies with BioClinicalBERT}
To make the synthetic letters more readable, clinically sound, and privacy-protective, different masking strategies are tested based on the following principles.
\begin{enumerate}
\item \textbf{Preserve Annotated Entities: }The manually annotated entities should not be masked to retain the clinical knowledge and context.
\item \textbf{Preserve Extracted Structures: }Tokens that are part of the document structure should be preserved as templates for clinical letters.
\item \textbf{Mask Detected Private Information: }This is helpful in de-identification. Although the dataset we use is de-identified, this approach may be useful when this system is deployed with real-world data.
\item \textbf{Preserve Medical Terminology: }It still aims to retain clinical knowledge, as some diseases and treatments were not manually annotated.
\item \textbf{Preserve Non-Private Numbers: }Certain numbers, such as drug dosage or heart rates, are indispensable for clinical diagnosis and treatment. However, only non-private numbers should be retained, while private information (such as phone numbers, ages, postal codes, dates, and email addresses) should be masked.
\item \textbf{Preserve Punctuation: } Punctuation marks such as periods (`.') and underscores (`\textunderscore\textunderscore\textunderscore') should not be masked, as they clarify the sentence boundaries and make the synthetic letters more coherent \cite{lamprou2022role}.
\item \textbf{Retain Special Patterns in Samples: }Tokens that match specific patterns (e.g. `Vitamin C \textasciicircum 1000 mg', `Ibuprofen \textgreater{} 200 mg', etc) should be retained, as they may contain important clinical details. These patterns are summarised by analysing raw sample letters.
\end{enumerate} 








Based on the principles above, different masking strategies were experimented with:
\begin{enumerate}
\item \textbf{Mask Randomly: }Tokens that can be masked are selected randomly from the text. We experimented with \textit{masking ratios} ranging from 0\% to 100\% in 10\% increments. This approach helps to understand how the number of masked tokens influences the quality of synthetic letters, and provides a baseline for other masking strategies.

\item \textbf{Mask Based on POS Tagging: } We experimented with different configurations in this section, such as masking only \textbf{nouns}, only \textbf{verbs}, etc. It is helpful to understand how POS influences the models' context understanding. Similar to the random masking approach, We selected the tokens based on their POS configuration and masked them in 10\% increments from 0\% to 100\%.





\item \textbf{Mask Stopwords: }Stopwords generally contribute little to the text's main idea. Masking stopwords serves two purposes: reducing the \textit{noise} for model understanding and increasing the \textit{variety} of synthetic text by predicting these words. Moreover, they do not influence crucial clinical information. This approach is highly similar to the one used in `Mask Based on POS Tagging'. The only difference is the criteria for selecting tokens. Specifically, tokens are selected based on whether they are stopwords rather than on their POS. `\textbf{NLTK}' library is used for detecting stopwords in the text.
\item \textbf{Hybrid Masking Using Different Ratio Settings: }After employing the aforementioned masking strategies, we observed the influence of these elements. Additionally, we experimented with their \textit{combinations} at different masking ratios based on the outcomes, such as masking 50\% nouns and 50\% stopwords simultaneously.
\end{enumerate}

\subsubsection{Determining Variable-Length-Chunk Size with Bio\_ClinicalBERT}
\label{subsec: Determining Chunk Size with BioClinicalBERT}


 As mentioned in Section \ref{sec: Pre-Processing}, we utilise two parameters in our chunk segment procedure: `max\_lines' and `max\_tokens'. `max\_lines' represents the desired length of each chunk, while `max\_tokens' is related to the computing resources and model limitations. These two parameters determine the final length of each chunk together. Although most models we used have a limit of 512 tokens (except for the Longformer, which can process up to 4096 tokens), we set 256 as the value for \textbf{`max\_tokens’} due to the computing resources constraints. 

As for `max\_lines', we experimented with values starting from 10 lines, increasing by 10 lines each time, and calculated the average tokens for each chunk. Once the token growth began to slow, we refined the search by using \textbf{finer increments}. Finally, we selected the number of lines at which the average tokens per chunk stopped growing. This is because more lines in each chunk provide more information for the model to predict masked tokens. However, if the chunk length reaches a critical threshold, it indicates that the primary limitation is  `max\_tokens', not `max\_lines'. Continuing to increase `max\_lines' would lead to additional computational overhead, as the system would have to repeatedly check whether adding the next sentence meets the required line count.

    \subsection{Evaluation Methods}
    \label{sec: Evaluation Methods}
\begin{figure}[htbp]
  \centering
  \includegraphics[width=0.83\linewidth]{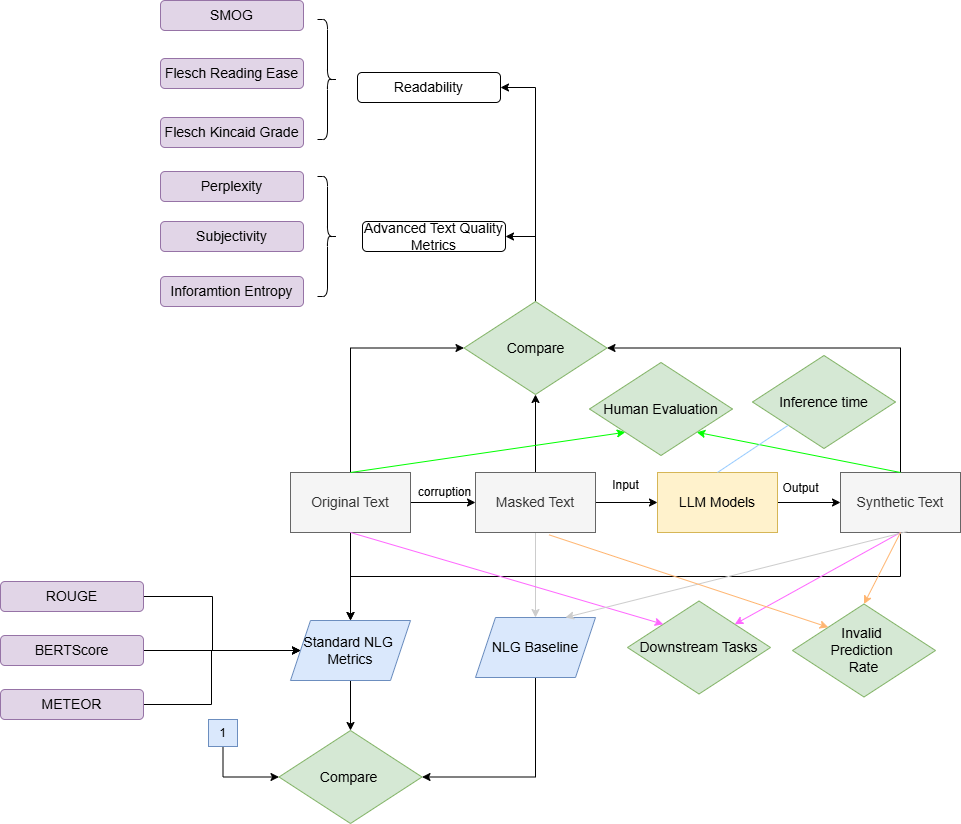}
  \caption{Evaluation Pipeline}           
  \label{fig: Evaluation Pipeline}  
\end{figure}

Both quantitative and qualitative methods will be used to evaluate the performance. Additionally, a downstream task (NER) is employed to assess whether the synthetic clinical letters can replace the original raw data. The evaluation methods pipeline is illustrated in Fig \ref{fig: Evaluation Pipeline}.
    
    \subsubsection{Quantitative Evaluation}
 
To comprehensively evaluate the quality of the synthetic letters, we used quantitative evaluation from multiple dimensions, including the model's \textbf{inference} performance, the \textbf{readability} of the synthetic letters, and their \textbf{similarity} to the raw data. The specific metrics are listed below.

\textbf{Standard NLG Metrics}
 It covers standard NLG evaluation methods such as ROUGE, BERTScore, and METEOR. 
 ROUGE measures literal similarity, BERTScore evaluates semantic similarity, and METEOR builds on ROUGE by taking synonyms and word order into account. It provides a more comprehensive evaluation of synthetic text \cite{banerjee-lavie-2005-meteor}. 

These evaluations will be performed by comparing synthetic text with the original text. Moreover, a baseline is calculated by comparing masked text to the original text. The evaluation score should exceed the baseline but remain below `1', ensuring it does not exactly replicate the original text.

\textbf{Readability Metrics}
 To evaluate the readability, we calculated \textbf{SMOG}, \textbf{Flesch Reading Ease}, and \textbf{Flesch-Kincaid Grade Level}. Given our clinical focus, we prioritise SMOG as the primary readability metric, with Flesch Reading Ease and Flesch-Kincaid Grade Level as reference standards. In this analysis, we will compare the readability metrics of the synthetic text with those of the original and masked texts. The evaluation results should closely approximate the original text's metrics. Significant differences may suggest that the model can not preserve semantic coherence and readability adequately.

\textbf{Advanced Text Quality Metrics}
 In this part, we calculated the \textbf{perplexity}, \textbf{subjectivity}, and \textbf{information entropy}. We want the synthetic letters to be useful in training clinical models. Therefore, perplexity should not be far away from the value of the original letters. As for subjectivity and information entropy, we expect the synthetic letters to be both subjective and informative.

\textbf{Invalid Prediction Rate}
 We calculated the invalid prediction rate for each generation configuration. This ratio is determined by dividing the number of invalid predictions (such as punctuation marks or subwords) by the total number of masked words that need to be predicted. We expect the model to generate more meaningful words. Since punctuation marks are not masked, the model should avoid generating too many non-words. This metric can provide insights into the model’s inference capability.

\textbf{Inference Time}
 The inference time for each generation configuration across the whole dataset (204 clinical letters) was recorded. Shorter inference times indicate lower computational resource consumption.  When this system is deployed on large datasets, it is expected to save both time and computing resources.

    \subsubsection{Qualitative Evaluation}
 In the quantitative evaluation, we not only calculated the evaluation metrics for the entire dataset, but also recorded the results for each individual synthetic clinical letter. Interestingly, while some synthetic texts exhibited strong performance according to most metrics, they did not always appear satisfactory upon ``visual" inspection. Conversely, some synthetic letters with average metrics may appear more visually appealing.

Although human evaluation is the most reliable approach for evaluating clinical letters, it is limited by availability and cost. 
Therefore, combining qualitative and quantitative evaluations helps in identifying suitable quantitative metrics for assessing our model’s performance. Once identified, one of these metrics can be used as the \textbf{primary standard}, while the others serve as supporting indicators. As a workaround, we selected a small sample of representative clinical letters based on the evaluation results. Subsequently, we reviewed the outcomes to better understand how different generation methods impacted these results, while also evaluating their correspondence with the quantitative metrics.

    \subsubsection{Downstream NER Task}
 Beyond qualitative and quantitative evaluation, we can also apply synthetic clinical letters in a downstream NER task. This is helpful to further evaluate their quality and their potential to replace original ones in clinical research and model training.

 ScispaCy \footnote{\url{https://allenai.github.io/scispacy/}} and spaCy \footnote{\url{https://spacy.io/}} are used in this part. As shown in Fig \ref{fig: Training Process for spaCy NER Model}, they extract features from the text and learn the weights of each feature through neural networks. These weights are updated by comparing the loss between the predicted probabilities and actual labels. If a word does not belong to any label, it is classified as `O' (outside any entity). SpaCy initialises these weights randomly.  However, the version of ScispaCy we use, (`en\_ner\_bc5cdr\_md'), is specifically fine-tuned on the \textbf{BC5CDR} corpus. It focuses more on the entities `chemical' and `disease' while retaining the original general features. 

\begin{figure}[htbp]
  \centering
  \includegraphics[width=\textwidth]{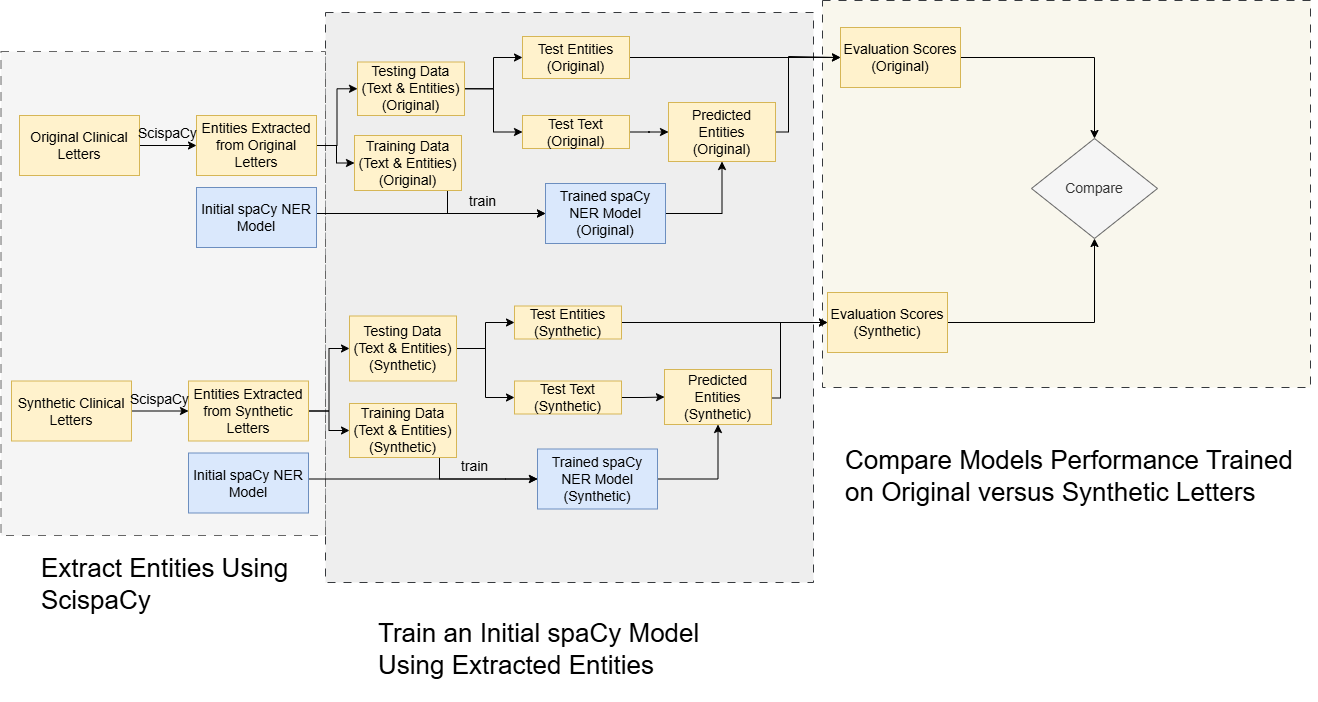}
  \caption{Workflow of Downstream NER Task}           
  \label{fig: Workflow of Downstream NER Task}  
\end{figure}

In this downstream NER task, as shown in Fig \ref{fig: Workflow of Downstream NER Task},  we initially extracted entities from letters using ScispaCy. Subsequently, these entities were used to train a base spaCy model. The trained model was then employed to extract entities from the testing set. Finally, we can compare these newly extracted entities with those originally extracted by ScispaCy, and the evaluation scores can be calculated. These steps were performed on both original clinical letters and synthetic letters, to assess whether the synthetic letters can potentially replace the original ones. 



    \subsection{Post-Processing}
    
\subsubsection{Filling in the blanks}
 As described in Section \ref{sec: Experimental Setup}, the dataset we used has been de-identified. All private information is replaced by three underscores `\_\_\_'. We hope that the synthetic clinical letters can maintain a certain degree of clinical integrity without revealing any private patient information. To address this, a post-processing step was added to the synthetic results. This process involves masking the three underscores (`\_\_\_') detected and using PLMs to predict the masked part again. For example, if the original text is `\_\_\_ caught a cold'. The post-processing result should ideally be `John caught a cold’ or `Patient caught a cold’.  Such synthetic clinical letters can better support clinical model training and teaching. 

In this part, we used Bio\_ClinicalBERT and BERT-base models. Although Bio\_ClinicalBERT is better at clinical information understanding, this issue is not directly related to clinical practice, so we used BERT-base for comparison.

\subsubsection{Spelling Correction}
 Since our data comes from real-world sources, it is inevitable that some words may be misspelled by doctors. These spelling errors can negatively impact the model's training process or hinder clinical practitioners' understanding of the synthetic clinical letters. Although some errors are masked and re-generated, our masking ratio is not always 100\%, so some incorrect words may still exist. A toolkit `TextBlob' \cite{textblob} is added to correct these errors. Specifically, it uses a rule-based approach that relies on a \textbf{built-in vocabulary library} to detect and correct misspellings.

    \subsection{Summary}
\begin{figure}[htbp]
  \centering
  \includegraphics[width=0.99\linewidth]{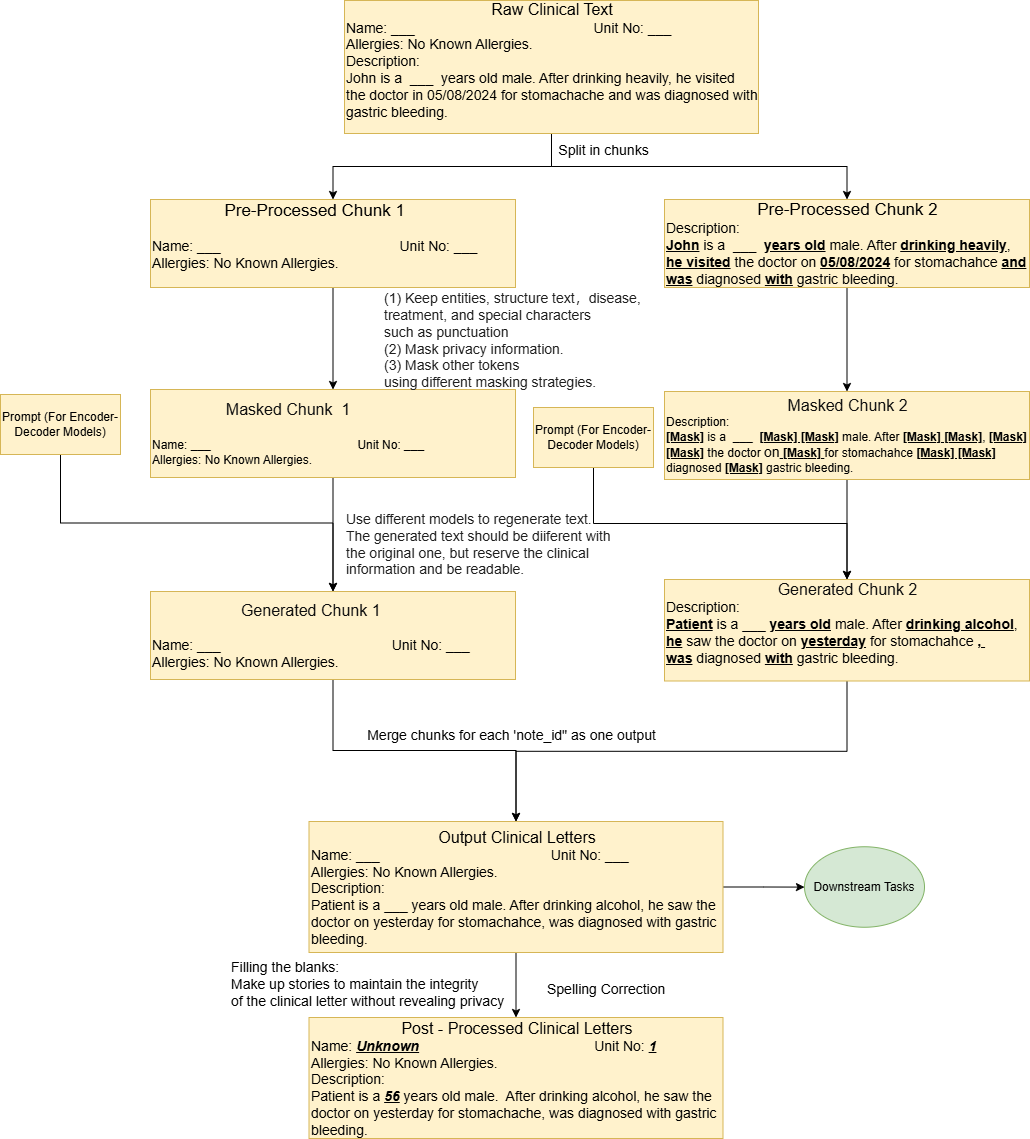}
  \caption{An Example of Masking and Generating}           

  \label{fig: An Example of Masking and Generating}  
\end{figure}

 In this section, we introduced the experimental design and subsequent implementation steps: from project requirement, data collection to environmental setup, pre-processing, masking and generating, post-processing, downstream NER task, and both qualitative and quantitative evaluation. An example of the entire process flow is shown in Figure \ref{fig: An Example of Masking and Generating}.

\section{Experimental Results and Analysis} 

    \subsection{Random Masking: Qualitative Results}
    \label{subsec: Qualitative Results of Models Comparison With Random Masking}
 We employed both the encoder-only and encoder-decoder models to mask and generate the data, yielding 
numerous interesting results for human evaluation. Given space constraints, only a simple example is provided here. Following the masking principles in Section \ref{sec: Clinical Letters Generation}, the eligible tokens were randomly selected for masking. Although the initial intention was to mask 50\% of tokens, the actual masking ratio was lower due to the requirement to preserve certain entities and structures.

\begin{figure}[htbp]
  \centering
  \includegraphics[width=0.8\linewidth]{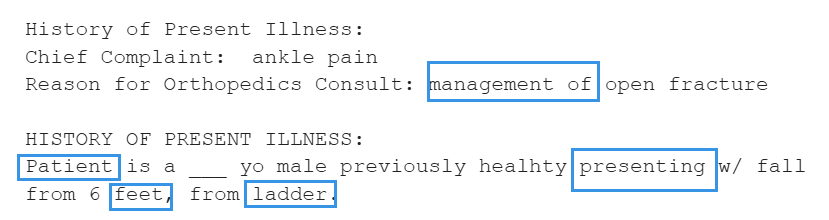}
  \caption{Original Unprocessed Example Sentence \cite{goldberger2000physiobank,johnson2024mimiciv,johnson2023mimic} (`note\_id': '10807423-DS-19') (The circled tokens will be masked)}           

  \label{fig: the original unprocessed example sentence}  
\end{figure}

\begin{table}[htbp]
\centering
\begin{tabular}{|c|c|c|c|}
\hline
\textbf{Entity} & \textbf{Start} & \textbf{End} & \textbf{Concept ID} \\ \hline
ankle pain & 411 & 421 & 247373008 \\ \hline
open fracture & 468 & 481 & 397181002 \\ \hline
fall & 571 & 575 & 161898004 \\ \hline
\end{tabular}
\caption{Annotated Entities Extracted from the Example Sentence (They should be preserved from masking)}
\label{Tab: Annotated Entities Extracted from the Example Sentence}
\end{table}

\subsubsection{Encoder-Only Models}
\begin{figure}[htbp]
  \centering
  \includegraphics[width=0.8\linewidth]{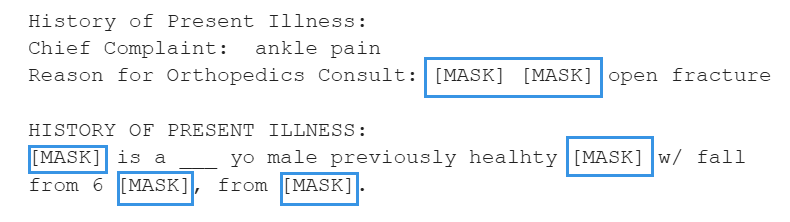}
  \caption{An Example of the Masked Sentence}           

  \label{fig: An Example of the Masked Sentence}  
\end{figure}

\begin{figure}[htbp]
  \centering
  \includegraphics[width=0.5\linewidth]{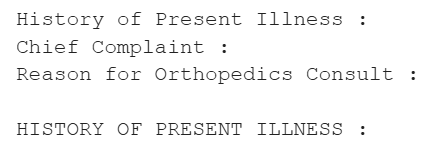}
  \caption{Extracted Structure from the Example Sentence}           

  \label{fig: Extracted Structure from the Example Sentence}  
\end{figure}

 The original sentence is displayed in Fig \ref{fig: the original unprocessed example sentence}. After the feature extraction, the resulting structure is shown in Fig \ref{fig: Extracted Structure from the Example Sentence}.
As detailed in Table \ref{Tab: Annotated Entities Extracted from the Example Sentence},  certain manually annotated entities are excluded from masking.
The output of this masking process can be seen in Fig \ref{fig: An Example of the Masked Sentence}. 


\begin{figure}[htbp]
  \centering
  \includegraphics[width=0.8\linewidth]{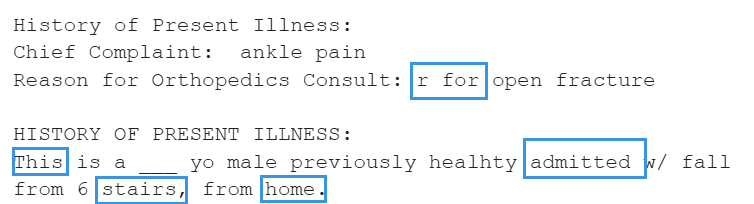}
  \caption{Example Sentence Generated by Bio\_ClinicalBERT}           

  \label{fig: Example Sentence Generated by BioClinicalBERT}  
\end{figure}    

The generated text using Bio\_ClinicalBERT is displayed in Fig \ref{fig: Example Sentence Generated by BioClinicalBERT}.  For 
`management of open fracture,' the model produced `r', which is commonly used to denote `right' in clinical contexts, showing a relevant and logical prediction. Furthermore, the model’s input `R ankle', despite not being in the figure due to space constraints, provided context for predicting `r' instead of `left.' Interestingly, the term `admitted' was generated even though it was not in the input, indicating the model's understanding of clinical context. Although the phrase `from 6 stairs, from home' differs significantly from the original one, it remains contextually appropriate.

Overall, Bio\_ClinicalBERT produced a clinically sound sentence, even though no tokens matched the original. In other examples, the predicted words may partially overlap with the original text. Nonetheless, this model effectively retains clinical information and introduces diversity without altering the text's meaning.

\begin{figure}[htbp]
  \centering
  \includegraphics[width=0.8\linewidth]{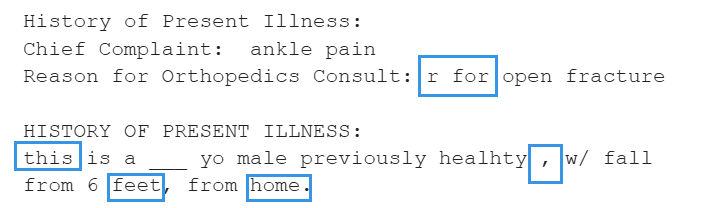}
  \caption{Example Sentence Generated by medicalai / ClinicalBERT}           

  \label{fig: Example Sentence Generated by medicalai/ClinicalBERT}  
\end{figure}

\begin{figure}[htbp]
  \centering
  \includegraphics[width=0.8\linewidth]{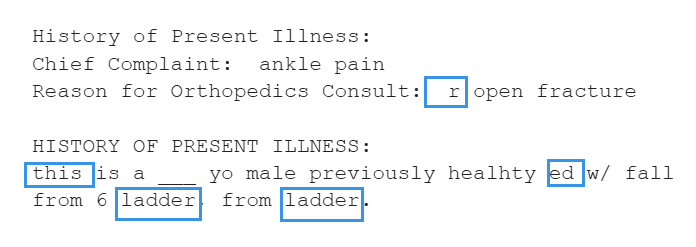}
  \caption{Example Sentence Generated by Clinical-Longformer}           

  \label{fig: Example Sentence Generated by Clinical-Longformer}  
\end{figure}

The results from medicalai/ClinicalBERT and Clinical-Longformer are shown in Fig \ref{fig: Example Sentence Generated by medicalai/ClinicalBERT} and Fig \ref{fig: Example Sentence Generated by Clinical-Longformer}. All three clinical-related models correctly predicted `r' from the input context. Medicalai/ClinicalBERT performs \textit{comparably} to Bio\_ClinicalBERT, despite adding an extra comma, which did not affect the text's clarity. 
However, Clinical-Longformer’s predictions, while understandable, were \textit{repetitive} and less satisfactory. Importantly, none of these three models altered the original meaning.

\begin{figure}[htbp]
  \centering
  \includegraphics[width=0.8\linewidth]{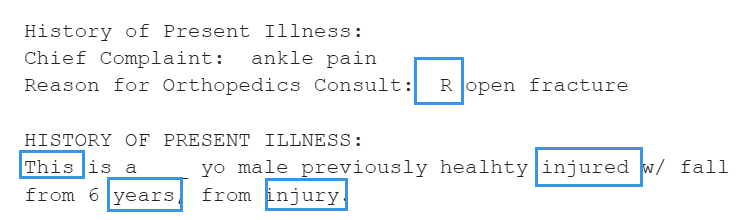}
  \caption{Example Sentence Generated by RoBERTa-base}           

  \label{fig: Example Sentence Generated by RoBERTa-base}  
\end{figure}  

The result generated by RoBERTa-base is shown in Fig \ref{fig: Example Sentence Generated by RoBERTa-base}. While the generated text initially seems reasonable, the predicted word `years' shifts the focus to a temporal context, which was not intended. 
This is likely because RoBERTa is pre-trained on a general corpus and lacks sufficient clinical knowledge for accurate text generation, or it could simply be a coincidence based on this specific sentence, where RoBERTa-base inferred ‘years’ from its training data.

\subsubsection{Decoder-Only GPT-4o}

\begin{figure}[htbp]
  \centering
  \includegraphics[width=0.8\linewidth]{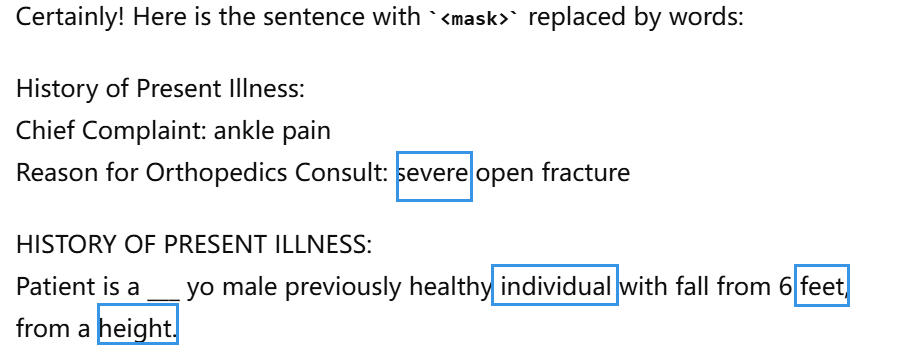}
  \caption{Example Sentence Generated by GPT-4o}           

  \label{fig: Example Sentence Generated by GPT-4o}  
\end{figure}

Additionally, \textbf{GPT-4o} was used for comparison, with the prompt ``Replace `\texttt{<mask>}' with words in the following sentence: ''. The results, shown in Fig \ref{fig: Example Sentence Generated by GPT-4o}, are satisfactory. As discussed in Section \ref{sec: Generative Language Models}, decoder-only models excel in few-shot learning \cite{wu2024large}, which is confirmed by this experiment. However, its performance may decline with long clinical letters \cite{amin-nejad-etal-2020-exploring}. 


\subsubsection{Encoder-Decoder Models}
\begin{figure}[htbp]
  \centering
  \includegraphics[width=0.8\linewidth]{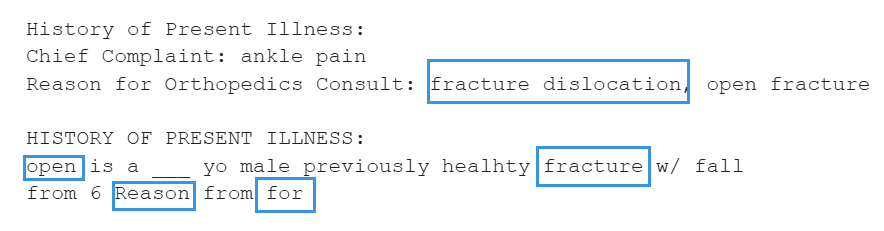}
  \caption{Example Sentence Generated by T5-base}           

  \label{fig: Example Sentence Generated by T5-base}  
\end{figure}

\begin{figure}[htbp]
  \centering
  \includegraphics[width=0.8\linewidth]{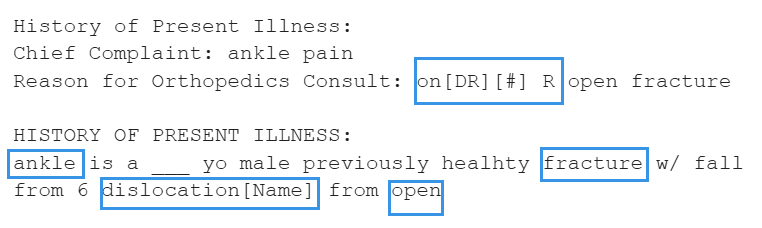}
  \caption{Example Sentence Generated by Clinical-T5-Base}           

  \label{fig: Example Sentence Generated by Clinical-T5-Base}  
\end{figure}

\begin{figure}[htbp]
  \centering
  \includegraphics[width=0.8\linewidth]{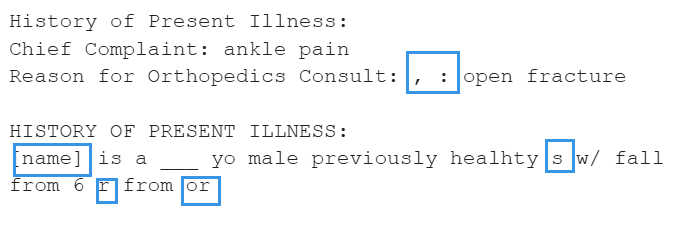}
  \caption{Example Sentence Generated by Clinical-T5-Scratch}           

  \label{fig: Example Sentence Generated by Clinical-T5-Scratch}  
\end{figure}

\begin{figure}[htbp]
  \centering
  \includegraphics[width=0.8\linewidth]{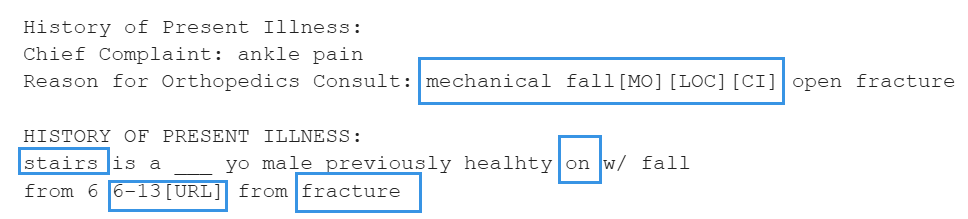}
  \caption{Example Sentence Generated by Clinical-T5-Sci}           

  \label{fig: Example Sentence Generated by Clinical-T5-Sci}  
\end{figure}

 To further evaluate different PLMs in generating synthetic letters, we tested the T5 Family models. The generated results for the same sentence are shown in Fig \ref{fig: Example Sentence Generated by T5-base}, Fig \ref{fig: Example Sentence Generated by Clinical-T5-Base}, Fig \ref{fig: Example Sentence Generated by Clinical-T5-Scratch}, and Fig \ref{fig: Example Sentence Generated by Clinical-T5-Sci}. 

T5-base performs the best among these tested models. However, the results are still \textbf{not fully rational}, as it generated `open is a \_\_\_ yo male'.
The other three models tend to use de-identification (\textbf{DEID}) tags to replace the masked words, as these tags are part of their corpora. Furthermore, the T5 family models may predict multiple words for each token, aligning with findings in Section \ref{sec: Generative Language Models}

All these four T5 family models perform \textbf{worse than the encoder-only} models. This is consistent with the findings from \cite{DBLP:journals/corr/abs-2405-12630} that Masked Language Modelling (MLM) models significantly outperform Causal Language Modeling (CLM) models in medical datasets.

    \subsection{Random Masking: Quantitative Results}
    \label{Subsec: Quantitative Results}
\subsubsection{Sentence-Level Quantitative Results: Encoder-Only Models}
 We first calculated representative quantitative metrics at the sentence level, matching the sample sentence used in Subsection \ref{subsec: Qualitative Results of Models Comparison With Random Masking}. This approach allows for a better integration of quantitative and qualitative evaluations.
Although SMOG is typically suited for medical datasets, it is less appropriate for sentence-level analysis, so the Flesch Reading Ease was used here. The results are shown in table \ref{Encoder-Only Models Comparison at the Sentence Level}.

\begin{table}[htbp]
\centering
\begin{tabular}{|c|p{2.4cm}|p{2.4cm}|p{2.4cm}|p{2.4cm}|}
\hline
 & \multicolumn{4}{|c|}{\textbf{Model Evaluation}} \\
\cline{2-5}
                & RoBERTa-base & medicalai / ClinicalBERT & Clinical-Longformer & Bio \_ ClinicalBERT \\ 
\hline
\multicolumn{5}{|c|}{\textbf{ROUGE-1}} \\
\hline
Generation Performance      & 86.54   & 88.46        & 89.52               & 84.91   \\
Baseline        & 84.91   & 84.91        & 84.91               & 84.91   \\
\hline
\multicolumn{5}{|c|}{\textbf{ROUGE-2}} \\
\hline
Generation Performance      & 74.51   & 78.43        & 79.61               & 73.08   \\
Baseline        & 73.08   & 73.08        & 73.08               & 73.08   \\
\hline
\multicolumn{5}{|c|}{\textbf{ROUGE-L}} \\
\hline
Generation Performance     & 86.54   & 88.46        & 89.52               & 84.91   \\
Baseline        & 84.91   & 84.91        & 84.91               & 84.91   \\
\hline
\multicolumn{5}{|c|}{\textbf{BERTScore F1}} \\
\hline
Generation Performance      & 0.81    & 0.83         & 0.84                & 0.85    \\
Baseline        & 0.79    & 0.65         & 0.79                & 0.65    \\
\hline
\multicolumn{5}{|c|}{\textbf{METEOR}} \\
\hline
Generation Performance      & 0.87    & 0.88         & 0.90                & 0.86    \\
Baseline        & 0.85    & 0.85         & 0.85                & 0.85    \\
\hline
\multicolumn{5}{|c|}{\textbf{Flesch Reading Ease}} \\
\hline
Generation Performance     & 10.24   & 18.70        & 9.22                & 16.67   \\
Baseline (Original)      & 8.21    & 8.21         & 8.21                & 8.21    \\
Baseline (Mask)       & 16.67     & 16.67     & 16.67                & 16.67  \\
\hline
\end{tabular}
\caption{Encoder-Only Models Comparison at the Sentence Level (The `Baseline' without annotations was calculated by comparing masked text to the original text)}
\label{Encoder-Only Models Comparison at the Sentence Level}
\end{table}

Our objective is to generate letters that differ from the original while maintaining clinical semantics and structure. Thus, high ROUGE scores are not desired, as they indicate significant \textbf{word/string} overlap. BERTScore is particularly useful for assessing \textbf{semantic} similarity, while METEOR offers a comprehensive evaluation considering word forms and synonyms theoretically. Flesch Reading Ease, on the other hand, provides a direct measure of textual \textbf{readability}.

We observed that clinical-related encoder-only models generally outperform RoBERTa-base in qualitative evaluation  (see Subsection \ref{subsec: Qualitative Results of Models Comparison With Random Masking}). 
However, from the quantitative perspective, RoBERTa-base shows mediocre performance across most metrics except for BERTScore. In contrast, Bio\_ClinicalBERT, despite no word overlap in this sample sentence, achieves a reasonable clinical context and the highest BERTScore among the models. Both Medicalai/Clinical BERT and Bio\_ClinicalBERT excel in Flesch Reading Ease, likely because they tend to predict tokens with fewer syllables words that preserve the original meaning.

Surprisingly, while METEOR is designed to closely reflect human evaluation, BERTScore appears to be more consistent with our evaluation criteria. This trend was observed in other sample texts as well. \textit{Synthetic texts with higher BERTScore and lower ROUGE scores are more aligned with our objectives}. It is likely because BERTScore is calculated using word embeddings, which can capture deep semantic similarity more effectively. All evaluation results \textit{meet or exceed the baseline, affirming the effectiveness of these four encoder-only models} in generating clinical letters. 

\subsubsection{Sentence-Level Quantitative Results: Encoder-Decoder Models}
 The evaluations for the encoder-decoder models, as shown in Table \ref{tab: Encoder-Decoder Models Comparison at the Sentence Level}, generally underperform on most 
metrics compared to encoder-only models, except for METEOR. Interestingly, while the Flesch Reading Ease scores suggest a minimal impact on readability, the BERTScores are significantly lower than the baseline, indicating major deviations from the original meaning. This is consistent with our qualitative observations that the outputs from encoder-decoder models are largely unintelligible. 


\begin{table}[htbp]
\centering
\begin{tabular}{|c|p{2.4cm}|p{2.4cm}|p{2.4cm}|p{2.4cm}|}
\hline
 & \multicolumn{4}{|c|}{\textbf{Model Evaluation}} \\
\cline{2-5}
                & T5-base & Clinical-T5-base & Clinical-T5-Scratch & Clinical-T5-Sci \\ 
\hline
\multicolumn{5}{|c|}{\textbf{ROUGE-1}} \\
\hline
Generation Performance      & 86.79   & 85.19        & 87.38               & 80.36   \\
Baseline        & 73.77   & 73.77        & 73.77               & 73.77   \\
\hline
\multicolumn{5}{|c|}{\textbf{ROUGE-2}} \\
\hline
Generation Performance      & 75.00   & 71.70        & 75.25               & 69.09   \\
Baseline        & 63.33   & 63.33        & 63.33               & 63.33   \\
\hline
\multicolumn{5}{|c|}{\textbf{ROUGE-L}} \\
\hline
Generation Performance      & 84.91   & 83.33        & 87.38               & 80.36   \\
Baseline        & 73.77   & 73.77        & 73.77               & 73.77   \\
\hline
\multicolumn{5}{|c|}{\textbf{BERTScore F1}} \\
\hline
Generation Performance     & 0.44    & 0.40         & 0.45                & 0.40    \\
Baseline        & 0.50    & 0.50         & 0.50                & 0.50    \\
\hline
\multicolumn{5}{|c|}{\textbf{METEOR}} \\
\hline
Generation Performance      & 0.85    & 0.83         & 0.83                & 0.82    \\
Baseline        & 0.85    & 0.85         & 0.85                & 0.85    \\
\hline
\multicolumn{5}{|c|}{\textbf{Flesch Reading Ease}} \\
\hline
Generation Performance      & 8.21    & 8.21         & 19.71               & 8.21    \\
Baseline (Original)         & 8.21    & 8.21         & 8.21                & 8.21    \\
Baseline (Mask)       & 8.21     & 8.21     & 8.21                & 8.21  \\
\hline
\end{tabular}
\caption{Encoder-Decoder Models Comparison at the Sentence Level (The Baseline without annotations was calculated by comparing masked text to the original text)}
\label{tab: Encoder-Decoder Models Comparison at the Sentence Level}
\end{table}

Collectively, the quantitative and qualitative results demonstrate that \textit{\textbf{encoder-decoder models are not well-suited for generating clinical letters}}, as they fail to preserve the original narratives. These results also support the validity of using BERTScore as the primary evaluation metric, with other metrics serving as supplementary references. We also tested this on the \textbf{entire dataset}, which produced \textit{consistent} results.


\subsubsection{Quantitative Results on the Full Dataset: Encoder-Only Models}
 Based on the findings above, we expect a higher BERTScore and a lower ROUGE Score. We used the 0.4 masking ratio to illustrate the model comparison on the full dataset in Table \ref{tab:encoder_only_models_comparison_full_dataset}. The other masking ratios show similar trends. 
 Surprisingly, all encoder-only models this time showed comparable results, which contradicts our hypothesis that ‘Clinical-related’ models would outperform base models. This suggests that \textit{training on the clinical dataset does not significantly impact the quality of synthetic letters}. This may be because most clinical-related tokens are preserved, with only the remaining tokens being eligible for masking. Consequently, the normal encoder-only models can effectively understand the context and predict appropriate words while preserving clinical information. This differs slightly from the sentence-level comparisons, likely because the evaluation of a single sentence cannot fully represent the overall results. Despite this, BERTScore as a primary evaluation metric remains useful, as the correspondence between qualitative and quantitative evaluation is consistent, whether at the sentence or dataset level.

\begin{table}[htbp]
\centering
\begin{tabular}{|c|p{2.4cm}|p{2.4cm}|p{2.4cm}|p{2.4cm}|}
\hline
 & \multicolumn{4}{|c|}{\textbf{Model Evaluation}} \\
\cline{2-5}
                & RoBERTa-base & medicalai / ClinicalBERT & Clinical-Longformer & Bio\_ ClinicalBERT \\ 
\hline
\multicolumn{5}{|c|}{\textbf{ROUGE-1}} \\
\hline
Generation Performance      & 92.98   & 93.63        & 94.66               & 93.18   \\
Baseline        & 85.64   & 85.44        & 85.64               & 85.61   \\
\hline
\multicolumn{5}{|c|}{\textbf{ROUGE-2}} \\
\hline
Generation Performance     & 86.10   & 87.42        & 89.50               & 86.50   \\
Baseline        & 74.96   & 74.64        & 74.96               & 74.92   \\
\hline
\multicolumn{5}{|c|}{\textbf{ROUGE-L}} \\
\hline
Generation Performance      & 92.54   & 93.22        & 94.38               & 92.71   \\
Baseline        & 85.64   & 85.44        & 85.64               & 85.61   \\
\hline
\multicolumn{5}{|c|}{\textbf{BERTScore F1}} \\
\hline
Generation Performance      & 0.91    & 0.90         & 0.92                & 0.90    \\
Baseline        & 0.82    & 0.63         & 0.82                & 0.63    \\
\hline
\end{tabular}
\caption{Encoder-Only Models Comparison on the Full Dataset with Masking Ratio 0.4 (The Baseline was calculated by comparing masked text to the original text)}
\label{tab:encoder_only_models_comparison_full_dataset}
\end{table}

We will now explore how different \textit{masking ratios} affect the quality of synthetic clinical letters. For each model, we generated data with masking ratios from 0.0 to 1.0, in increments of 0.1 (the masking ratios here refer only to the eligible tokens, as described in Subsection \ref{subsec: Different Masking Strategies with BioClinicalBERT}, and do not represent the actual overall masking ratio). Due to space limitations, we will present only the results for Bio\_ClinicalBERT with a 0.2 increment here. 

Table \ref{tab: Classical NLP Metrics Across Different Masking Ratios} shows that the higher masking ratio, the lower the similarity (metrics' scores) will be. As we expected, all evaluation values are higher than the baseline, but still below `1'. This means the model can understand the clinical context and generate understandable text. It is surprising that with the masking ratio of 1.0, BERTScore increased from the baseline (0.29) to 0.63. Although this score is not very high, it still reflects that Bio\_ClinicalBERT can generate clinical text effectively.

\begin{table}[htbp]
\centering
\begin{tabular}{|c|c|c|c|c|c|c|}
\hline
\textbf{Bio\_ClinicalBERT} & \multicolumn{6}{|c|}{\textbf{Masking Ratio}} \\
\cline{2-7}
                & \textbf{1.0} & \textbf{0.8} & \textbf{0.6} & \textbf{0.4} & \textbf{0.2} & \textbf{0.0} \\ 
\hline
\multicolumn{7}{|c|}{\textbf{ROUGE-1}} \\
\hline
Generation Performance      & 76.28   & 83.75   & 88.91   & 93.18   & 96.76   & 99.51   \\
\hline
Baseline        & 64.05   & 71.56   & 78.56   & 85.61   & 92.63   & 99.22   \\
\hline
\multicolumn{7}{|c|}{\textbf{ROUGE-2}} \\
\hline
Generation Performance      & 62.60   & 70.77   & 78.81   & 86.50   & 93.42   & 99.02   \\
\hline
Baseline        & 51.72   & 57.88   & 65.38   & 74.92   & 86.27   & 98.61   \\
\hline
\multicolumn{7}{|c|}{\textbf{ROUGE-L}} \\
\hline
Generation Performance     & 74.33   & 81.69   & 87.71   & 92.71   & 96.65   & 99.50   \\
\hline
Baseline        & 64.05   & 71.56   & 78.56   & 85.61   & 92.63   & 99.22   \\
\hline
\multicolumn{7}{|c|}{\textbf{BERTScore}} \\
\hline
Generation Performance     & 0.63    & 0.75    & 0.83    & 0.90    & 0.95    & 0.99    \\
\hline
Baseline        & 0.29    & 0.39    & 0.50    & 0.63    & 0.79    & 0.98    \\
\hline
\multicolumn{7}{|c|}{\textbf{METEOR}} \\
\hline
Generation Performance     & 0.70    & 0.80    & 0.87    & 0.93    & 0.97    & 1.00    \\
\hline
Baseline        & 0.66    & 0.72    & 0.78    & 0.85    & 0.92    & 0.99    \\
\hline
\end{tabular}
\caption{Standard NLG Metrics Across Different Masking Ratios Using Bio\_ClinicalBERT (The Baseline was calculated by comparing masked text to the original text)}
\label{tab: Classical NLP Metrics Across Different Masking Ratios}
\end{table}

In Table \ref{tab:readability_metrics_masking_ratio}, we calculated three \textbf{readability} metrics, which were mentioned in Section \ref{sec: Evaluation Methods}. All these metrics have not shown significant changes from the original ones. However, it is strange that the SMOG and Flesh-Kincaid Grade are not always between the original baseline and mask baseline. When the masking ratio is high, the evaluation values even fall below both the masking and the original baseline. This may be because a \textit{higher masking ratio leads to a lower valid prediction rate}. If the predicted words include many spaces or punctuation marks, the readability will decrease obviously. 

\begin{table}[htbp]
\centering
\begin{tabular}{|c|c|c|c|c|c|c|}
\hline
\textbf{Bio\_ClinicalBERT} & \multicolumn{6}{|c|}{\textbf{Masking Ratio}} \\
\cline{2-7}
                & \textbf{1.0} & \textbf{0.8} & \textbf{0.6} & \textbf{0.4} & \textbf{0.2} & \textbf{0.0} \\ 
\hline
\multicolumn{7}{|c|}{\textbf{SMOG}} \\
\hline
Generation Performance      & 8.91 & 9.18 & 9.50 & 9.79 & 10.00 & 10.13 \\
Baseline (Original)      & 10.16 & 10.15 & 10.15 & 10.15 & 10.15 & 10.15 \\
Baseline (Mask)   & 9.04 & 9.29 & 9.52 & 9.74 & 9.95 & 10.13 \\
\hline
\multicolumn{7}{|c|}{\textbf{Flesch Reading Ease}} \\
\hline
Generation Performance     & 63.77 & 63.44 & 61.41 & 59.54 & 58.06 & 57.02 \\
Baseline (Original)      & 56.85 & 56.87 & 56.87 & 56.87 & 56.87 & 56.87 \\
Baseline (Mask)   & 70.11 & 67.39 & 64.75 & 62.15 & 59.62 & 57.13 \\
\hline
\multicolumn{7}{|c|}{\textbf{Flesch-Kincaid Grade}} \\
\hline
Generation Performance     & 7.32 & 7.70 & 8.24 & 8.66 & 9.01 & 9.22 \\
Baseline (Original)      & 9.26 & 9.26 & 9.26 & 9.26 & 9.26 & 9.26 \\
Baseline (Mask)   & 7.41 & 7.79 & 8.16 & 8.52 & 8.87 & 9.22 \\
\hline
\end{tabular}
\caption{Readability Metrics Across Different Masking Ratios Using Bio\_ClinicalBERT (The Baseline without annotations was calculated by comparing masked text to the original text)}
\label{tab:readability_metrics_masking_ratio}
\end{table}

In Table \ref{tab: Comparison of Advanced Text Quality Metrics Across Different Masking Ratios}, considering the perplexity, the masking baseline is very high, while the values for synthetic letters are close to the original ones. This indicates that the synthetic letters are useful for training clinical models. For information entropy, regardless of the masking ratio, it can \textit{effectively preserve the amount of information}. As for subjectivity, since all the values are close, we don’t need to worry that the synthetic letters will be biased.

\begin{table}[htbp]
\centering
\begin{tabular}{|c|c|c|c|c|c|c|}
\hline
\textbf{Bio\_ClinicalBERT} & \multicolumn{6}{|c|}{\textbf{Masking Ratio}} \\
\cline{2-7}
                & \textbf{1.0} & \textbf{0.8} & \textbf{0.6} & \textbf{0.4} & \textbf{0.2} & \textbf{0.0} \\ 
\hline
\multicolumn{7}{|c|}{\textbf{Perplexity}} \\
\hline
Generation Performance      & 2.24 & 2.32 & 2.31 & 2.30 & 2.29 & 2.29 \\
Baseline (Original)      & 2.22 & 2.28 & 2.28 & 2.28 & 2.28 & 2.28 \\
Baseline (Mask)   & 250.37 & 65.42 & 24.29 & 8.95 & 4.03 & 2.39 \\
\hline
\multicolumn{7}{|c|}{\textbf{Information Entropy}} \\
\hline
Generation Performance      & 5.46 & 5.80 & 5.92 & 5.96 & 5.98 & 5.98 \\
Baseline (Original)      & 5.98 & 5.98 & 5.98 & 5.98 & 5.98 & 5.98 \\
Baseline (Mask)   & 4.51 & 4.93 & 5.29 & 5.60 & 5.85 & 5.97 \\
\hline
\multicolumn{7}{|c|}{\textbf{Subjectivity}} \\
\hline
Generation Performance      & 0.32 & 0.32 & 0.32 & 0.32 & 0.33 & 0.33 \\
Baseline (Original)     & 0.33 & 0.33 & 0.33 & 0.33 & 0.33 & 0.33 \\
Baseline (Mask)   & 0.41 & 0.39 & 0.38 & 0.37 & 0.35 & 0.33 \\
\hline
\end{tabular}
\caption{Advanced Text Quality Metrics Across Different Masking Ratios Using Bio\_ClinicalBERT (The Baseline without annotations was calculated by comparing masked text to the original text)}
\label{tab: Comparison of Advanced Text Quality Metrics Across Different Masking Ratios}
\end{table}

As shown in Table \ref{tab: Inference Time and Invalid Prediction Rate Across Different Masking Ratios}, \textbf{inference time} for the entire dataset consistently ranges between 3 to 4 hours. However, it decreases with either very high or very low masking ratios. A mid-range masking ratio of approximately 0.6 results in longer inference times, likely because lower ratios reduce the number of masked tokens to process, while higher ratios provide less context, reducing the computational load. This lack of effective context also increases the invalid prediction rate. Conversely, with a masking ratio of `0', even a small number of prediction errors can significantly impact the overall accuracy due to the few masked tokens.
\begin{table}[htbp]
\centering
\begin{tabular}{|c|c|c|c|c|c|c|}
\hline
\textbf{Masking Ratio} & \textbf{1.0} & \textbf{0.8} & \textbf{0.6} & \textbf{0.4} & \textbf{0.2} & \textbf{0.0} \\ 
\hline
\textbf{Inference Time}      & 3:12:05 & 3:28:56 & 3:33:26 & 3:25:16 & 3:13:26 & 3:01:11 \\
\hline
\textbf{Invalid Prediction Rate}     & 0.72 & 0.47 & 0.34 & 0.28 & 0.25 & 0.37 \\
\hline
\end{tabular}
\caption{Inference Time and Invalid Prediction Rate Across Different Masking Ratios Using Bio\_ClinicalBERT}
\label{tab: Inference Time and Invalid Prediction Rate Across Different Masking Ratios}
\end{table}









    \subsection{Variable-Length Chunk Segmentation}
 As mentioned in Subsection \ref{subsec: Determining Chunk Size with BioClinicalBERT}, we set `max\_lines' as a variable and the `max\_tokens' equal to 256. A series of increasing `max\_lines' were tested until the average tokens per chunk reached a peak. 
 We initially did this on a small sample (7 letters). The results are shown in Table \ref{tab: Comparison for different Chunk Size}.

We can see that the average tokens per chunk reach a peak as the `max\_lines' parameter increases to 41. Similarly, inference time decreases as 'max\_lines' increases up to 41, but starts rising again once it exceeds this value. This experiment was also carried out on small samples of 10 and 30 letters. All of them showed the same trend. However, the inference time here may only reflect an overall trend, not exact results, as it is influenced by many factors, not only the chunk size but also the internet speed, etc.

\begin{table}[htbp]
\centering
\resizebox{\textwidth}{!}{  
\begin{tabular}{|p{1.5cm}|c|c|c|c|c|c|c|c|c|c|c|}
\hline
\textbf{max\_lines} & \textbf{10} & \textbf{20} & \textbf{30} & \textbf{35}  & \textbf{40} & \textbf{41} & \textbf{42} & \textbf{45} & \textbf{50} \\ 
\hline
\textbf{Inference Time (min)}  & 13:47 & 8:10 & 6:44 & 5:24 & 5:10 & 5:01 &  5:12 & 5:54  & 6:05 \\
\hline
\textbf{Average Tokens Per Chunk}     & 51.59 & 90.23 & 131.26 & 136.55  & 144.34  & 146.43 & 146.43 & 146.43 & 146.43 \\
\hline
\end{tabular}
}
\caption{Comparison for different Chunk Size}
\label{tab: Comparison for different Chunk Size}
\end{table}


    \subsection{Other Masking Strategies Using Bio\_ClinicalBERT} 
\begin{figure}[htbp]
  \centering
  \includegraphics[width=0.8\linewidth]{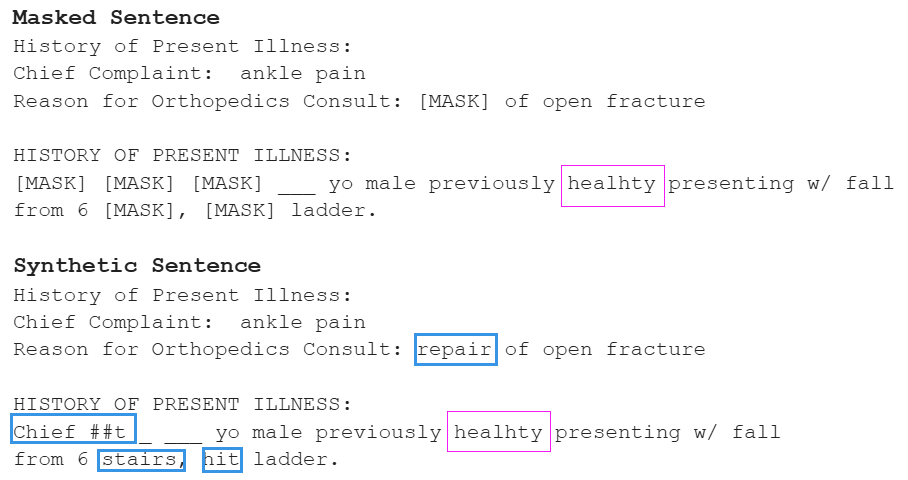}
  \caption{Example Sentence 1 with Different Masked Tokens}           

  \label{fig: Example Sentence 1 With Different Masking Positions}  
\end{figure} 

\begin{figure}[htbp]
  \centering
  \includegraphics[width=0.8\linewidth]{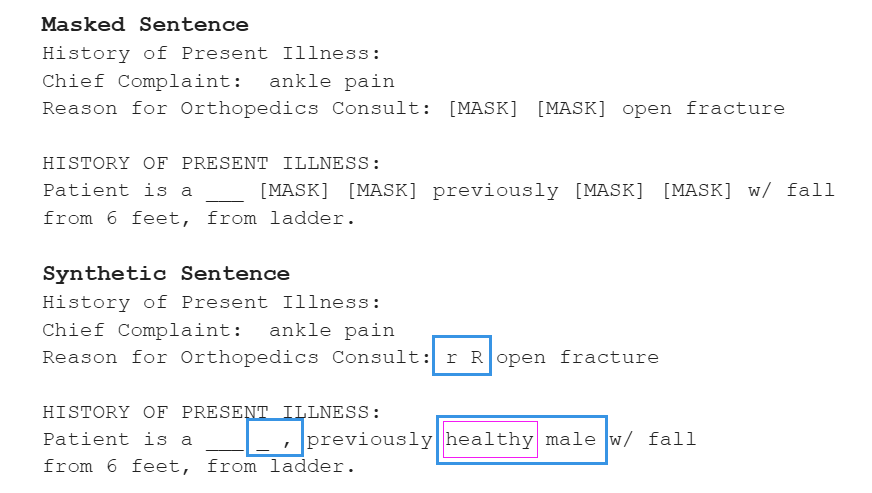}
  \caption{Example Sentence 2 with Different Masked Tokens}           

  \label{fig: Example Sentence 2 With Different Masking Positions}  
\end{figure}

 There is a random selection when masking tokens at certain ratios. Masking different types of tokens will lead to different results, as shown in Fig \ref{fig: Example Sentence 1 With Different Masking Positions} and Fig \ref{fig: Example Sentence 2 With Different Masking Positions}. This variability is understandable since the encoder-only models use bidirectional attention, as mentioned in Section \ref{sec: Generative Language Models}. \textit{These models need to predict the masked tokens based on the context}. Therefore, it is necessary to experiment with different masking strategies based on the types of tokens. We used \textbf{POS} tagging and \textbf{stopwords} to observe how these strategies influence the quality of synthetic letters.

As discussed in Subsection \ref{Subsec: Quantitative Results}, BERTScore should be the primary evaluation metric for our objective. Additionally, the invalid prediction rate is useful for assessing the model's ability to generate informative predictions, and ROUGE scores help evaluate literal diversity. Therefore, these quantitative metrics, calculated using different masking strategies, will be shown in this section. Similar to Subsection \ref{Subsec: Quantitative Results}, we experimented with different masking ratios calculated from the eligible tokens  (masked tokens divided by eligible tokens). The ratios are increased in increments of 0.1, ranging from 0.0 to 1.0. Due to space constraints, only metrics with increments of 0.2 will be shown here. A comparison with the same actual masking ratio (masked tokens divided by total tokens in the text) will also be presented in this subsection.

\subsubsection{Masking Only Nouns}
 Nouns often correspond to Personally Identifiable Information (\textbf{PII}), so masking nouns can serve as a verification step for de-identification. This experiment also helps identify which types of tokens for masking can introduce variations without significantly affecting the original clinical narratives.

As shown in Table \ref{tab: Quantitative Comparisons of Nouns Masking Ratios}, \textbf{the fewer nouns we mask, the better all these metrics perform}. This trend is consistent with random masking. When the noun masking ratio is 1.0, meaning all nouns are masked, BERTScore increases from a baseline of 0.70 to 0.89. This means the \textit{model predicted meaningful nouns}.  A similar trend is observed in the ROUGE scores. All evaluations are higher than the baseline but lower than `1'. However, ROUGE scores show a smaller improvement than BERTScore. This may be because the model generates synonyms or paraphrases that retain the original meaning. With the increase in nouns masking ratio, the BERTScore decreases significantly.

\begin{table}[htbp]
\centering
\begin{tabular}{|c|c|c|c|c|c|c|} 
\hline
\textbf{Bio\_ClinicalBERT} & \multicolumn{6}{|c|}{\textbf{Nouns Masking Ratio}} \\
\cline{2-7}
 & \textbf{1.0} & \textbf{0.8} & \textbf{0.6} & \textbf{0.4} & \textbf{0.2} & \textbf{0.0} \\ 
\hline
\multicolumn{7}{|c|}{\textbf{ROUGE-1}} \\
\hline
Generation Performance     & 93.29   & 95.16   & 96.48   & 97.62   & 98.66   & 99.51 \\ 
Baseline        & 88.13   & 90.79   & 92.98   & 95.19   & 97.39   & 99.22 \\
\hline
\multicolumn{7}{|c|}{\textbf{ROUGE-2}} \\
\hline
Generation Performance     & 86.71   & 90.29   & 92.84   & 95.12   & 97.28   & 99.02 \\
Baseline        & 78.32   & 82.92   & 86.79   & 90.82   & 95.01   & 98.61 \\
\hline
\multicolumn{7}{|c|}{\textbf{ROUGE-L}} \\
\hline
Generation Performance     & 93.00   & 94.96   & 96.35   & 97.56   & 98.64   & 99.50 \\
Baseline        & 88.13   & 90.79   & 92.98   & 95.19   & 97.39   & 99.22 \\
\hline
\multicolumn{7}{|c|}{\textbf{BERTScore}} \\
\hline
Generation Performance     & 0.89    & 0.92    & 0.94    & 0.96    & 0.98    & 0.99 \\
Baseline        & 0.70    & 0.76    & 0.81    & 0.86    & 0.92    & 0.98 \\
\hline
\multicolumn{7}{|c|}{\textbf{Invalid Prediction Rate}} \\
\hline
Generation Performance     & 0.37    & 0.34    & 0.33    & 0.32    & 0.32    & 0.37 \\ 
\hline
\end{tabular}
\caption{Quantitative Comparisons of Nouns Masking Ratios (The `Baseline' was calculated by comparing masked text to the original text)}
\label{tab: Quantitative Comparisons of Nouns Masking Ratios}
\end{table}

Therefore, to generate synthetic clinical letters that are distinguishable but still retain the original clinical information, we can only partially mask nouns (around a 0.8 masking ratio). It helps maintain balanced evaluation scores. If we mask all nouns, the quality of synthetic letters will decrease significantly.

\subsubsection{Masking Only Verbs}
 Masking only verbs also helps identify which token types are appropriate for masking to achieve our objective. While verbs are essential to describing clinical events, some can still be inferred from context. Therefore, \textit{masking verbs may have a slight effect on the quality of synthetic clinical letters, but it can also introduce some variation}.

Table \ref{tab: Quantitative Comparisons of Verb Masking Ratios} shows a similar trend for masking verbs as observed with other masking strategies in standard NLG metrics. However, it is surprising that as the masking ratio increases, both the invalid prediction rate and NLG metrics decrease. This phenomenon can be attributed to two main reasons. \textbf{First}, the model seems to prioritise predicting meaningful tokens (rather than punctuation, spaces, etc.) to generate coherent sentences. Contextual relevance is only considered after the sentence structure is complete. This may be due to the important role of verbs in sentences. \textbf{Second}, the original raw data may contain fewer verbs than nouns. Therefore, the number of actual masking tokens changes slightly when verbs are masked, making the model less sensitive to them. This is also reflected in BERTScore. If all verbs are masked, the BERTScore remains high at 0.95, whereas if all nouns are masked, the BERTScore drops to 0.89.

\begin{table}[htbp]
\centering
\begin{tabular}{|c|c|c|c|c|c|c|}
\hline
\textbf{Bio\_ClinicalBERT} & \multicolumn{6}{|c|}{\textbf{Verb Masking Ratio}} \\
\cline{2-7}
                & \textbf{1.0} & \textbf{0.8} & \textbf{0.6} & \textbf{0.4} & \textbf{0.2} & \textbf{0.0} \\ 
\hline
\multicolumn{7}{|c|}{\textbf{ROUGE-1}} \\
\hline
Generation Performance     & 96.48       & 97.38       & 97.97       & 98.54       & 99.08       & 99.51        \\
\hline
Baseline        & 94.11       & 95.50       & 96.48       & 97.50       & 98.48       & 99.22        \\
\hline
\multicolumn{7}{|c|}{\textbf{ROUGE-2}} \\
\hline
Generation Performance     & 92.79       & 94.63       & 95.84       & 97.04       & 98.15       & 99.02        \\
\hline
Baseline        & 88.53       & 91.26       & 93.18       & 95.19       & 97.14       & 98.61        \\
\hline
\multicolumn{7}{|c|}{\textbf{ROUGE-L}} \\
\hline
Generation Performance    & 96.37       & 97.31       & 97.92       & 98.51       & 99.07       & 99.50        \\
\hline
Baseline        & 94.11       & 96.48       & 96.48       & 97.50       & 98.48       & 99.22        \\
\hline
\multicolumn{7}{|c|}{\textbf{BERTScore}} \\
\hline
Generation Performance     & 0.95        & 0.97        & 0.97        & 0.98        & 0.99        & 0.99         \\
\hline
Baseline        & 0.82        & 0.86        & 0.89        & 0.92        & 0.95        & 0.98         \\
\hline
\multicolumn{7}{|c|}{\textbf{Invalid Prediction Rate}} \\
\hline
Generation Performance     & 0.31        & 0.31        & 0.32        & 0.32        & 0.33        & 0.37         \\
\hline
\end{tabular}
\caption{Quantitative Comparisons of Verb Masking Ratios (The `Baseline' was calculated by comparing masked text to the original text)}
\label{tab: Quantitative Comparisons of Verb Masking Ratios}
\end{table}

\subsubsection{Masking Only Stopwords}
 As mentioned in Subsection \ref{subsec: Different Masking Strategies with BioClinicalBERT}, masking stopwords aims to reduce noise for model understanding while introducing variation in synthetic clinical letters. Table \ref{tab: Quantitative Comparisons of Stopwords Masking Ratios} shows that \textit{masking only stopwords follows a similar trend to random masking}, where a higher masking ratio leads to lower ROUGE Score and BERTScore. Additionally, the Invalid Prediction Rate is at its lowest with a medium masking ratio. This is because higher masking ratios always result in more information loss. On the other hand, lower masking ratios lead to fewer tokens being masked, which makes small prediction errors more influential. The results show an overall low Invalid Prediction Rate and high BERTScore, indicating that \textit{stopwords have only a limited influence on the model's understanding of context}. This is not because the original raw letters contain very few stopwords. In fact, there are even more stopwords than nouns and verbs, as seen in sample texts.

\begin{table}[htbp]
\centering
\begin{tabular}{|c|c|c|c|c|c|c|}
\hline
\textbf{Bio\_ClinicalBERT} & \multicolumn{6}{|c|}{\textbf{Stopwords Masking Ratio}} \\
\cline{2-7}
                & \textbf{1.0} & \textbf{0.8} & \textbf{0.6} & \textbf{0.4} & \textbf{0.2} & \textbf{0.0} \\ 
\hline
\multicolumn{7}{|c|}{\textbf{ROUGE-1}} \\
\hline
Generation Performance      & 92.84       & 95.17       & 96.56       & 97.71       & 98.69       & 99.51        \\
\hline
Baseline        & 81.52       & 85.53       & 89.04       & 92.54       & 96.04       & 99.22        \\
\hline
\multicolumn{7}{|c|}{\textbf{ROUGE-2}} \\
\hline
Generation Performance     & 84.64       & 89.41       & 92.53       & 95.05       & 97.24       & 99.02        \\
\hline
Baseline        & 68.30       & 74.35       & 79.99       & 86.02       & 92.44       & 98.61        \\
\hline
\multicolumn{7}{|c|}{\textbf{ROUGE-L}} \\
\hline
Generation Performance     & 91.84       & 94.53       & 96.23       & 97.56       & 98.65       & 99.50        \\
\hline
Baseline        & 81.52       & 85.53       & 89.04       & 92.54       & 96.04       & 99.22        \\
\hline
\multicolumn{7}{|c|}{\textbf{BERTScore}} \\
\hline
Generation Performance      & 0.89        & 0.93        & 0.95        & 0.97        & 0.98        & 0.99         \\
\hline
Baseline        & 0.57        & 0.65        & 0.71        & 0.79        & 0.88        & 0.98         \\
\hline
\multicolumn{7}{|c|}{\textbf{Invalid Prediction Rate}} \\
\hline
Generation Performance     & 0.29        & 0.22        & 0.20        & 0.18        & 0.19        & 0.37         \\
\hline
\end{tabular}
\caption{Quantitative Comparisons of Stopwords Masking Ratios (The `Baseline' was calculated by comparing masked text to the original text)}
\label{tab: Quantitative Comparisons of Stopwords Masking Ratios}
\end{table}

\subsubsection{Comparison of Identical Actual Masking Ratios}
 To further observe how different masking strategies influence the generation of clinical letters, we compared the results using the same actual masking ratios but with different strategies. In other words, the number of masked tokens is fixed, so the only variable is \textit{the type of tokens being masked}. Table \ref{tab: Quantitative Comparisons of Masking Strategies} shows the results with a 0.04 actual masking ratio, and Table \ref{tab: Quantitative Comparisons of 0.1 overall} shows the results with a 0.1 actual masking ratio.

As we can see, masking only stopwords got the highest BERTScore and lowest invalid prediction rate. Therefore, stopwords have little influence on the overall meaning of the text, which is consistent with our earlier findings. Additionally, masking nouns and verbs perform worse than random masking. Therefore, if we want to preserve the original meaning, we cannot mask too many nouns and verbs.

\begin{table}[htbp]
\centering
\begin{tabular}{|c|p{2cm}|p{2cm}|p{2cm}|p{2cm}|}
\hline
\textbf{Bio\_ ClinicalBERT} & \multicolumn{4}{|c|}{\textbf{Masking Strategies}} \\
\cline{2-5}
                & \textbf{Noun Masking (0.4)} & \textbf{Stopwords Masking (0.2)} & \textbf{Verbs Masking (0.8)} & \textbf{Randomly Masking (0.1)} \\ 
\hline
\multicolumn{5}{|c|}{\textbf{ROUGE-1}} \\
\hline
Generation Performance      & 97.62       & 98.69       & 97.62       & 98.28        \\
\hline
Baseline        & 95.19       & 96.04       & 95.19       & 96.16        \\
\hline
\multicolumn{5}{|c|}{\textbf{ROUGE-2}} \\
\hline
Generation Performance     & 95.12       & 97.24       & 95.12       & 96.50        \\
\hline
Baseline        & 90.82       & 92.44       & 90.82       & 92.68        \\
\hline
\multicolumn{5}{|c|}{\textbf{ROUGE-L}} \\
\hline
Generation Performance      & 97.56       & 98.65       & 97.56       & 98.25        \\
\hline
Baseline        & 95.19       & 96.04       & 95.19       & 96.16        \\
\hline
\multicolumn{5}{|c|}{\textbf{BERTScore}} \\
\hline
Generation Performance      & 0.96        & 0.98        & 0.96        & 0.97         \\
\hline
Baseline        & 0.86        & 0.88        & 0.86        & 0.88         \\
\hline
\multicolumn{5}{|c|}{\textbf{Invalid Prediction Rate}} \\
\hline
Generation Performance      & 0.32        & 0.19        & 0.32        & 0.25         \\
\hline
\end{tabular}
\caption{Quantitative Comparison of Different Masking Strategies at a 0.04 Actual Masking Ratio (The `Baseline' was calculated by comparing masked text to the original text)}
\label{tab: Quantitative Comparisons of Masking Strategies}
\end{table}

\begin{table}[htbp]
\centering
\begin{tabular}{|c|p{2cm}|p{2cm}|p{2cm}|}
\hline
\textbf{Bio\_ClinicalBERT} & \textbf{Nouns Masking (1.0)} & \textbf{Stopwords Masking (0.6)} & \textbf{Random Masking (0.3)} \\ 
\hline
\multicolumn{4}{|c|}{\textbf{ROUGE-1}} \\
\hline
Generation Performance      & 93.29       & 96.56       & 95.10        \\
\hline
Baseline        & 88.13       & 89.04       & 89.16        \\
\hline
\multicolumn{4}{|c|}{\textbf{ROUGE-2}} \\
\hline
Generation Performance      & 86.71       & 92.53       & 90.17        \\
\hline
Baseline        & 78.32       & 79.99       & 80.44        \\
\hline
\multicolumn{4}{|c|}{\textbf{ROUGE-L}} \\
\hline
Generation Performance      & 93.00       & 96.23       & 94.86        \\
\hline
Baseline        & 88.13       & 89.04       & 89.16        \\
\hline
\multicolumn{4}{|c|}{\textbf{BERTScore}} \\
\hline
Generation Performance      & 0.89        & 0.95        & 0.93         \\
\hline
Baseline        & 0.70        & 0.71        & 0.71         \\
\hline
\multicolumn{4}{|c|}{\textbf{Invalid Prediction Rate}} \\
\hline
Generation Performance      & 0.37        & 0.20        & 0.26         \\
\hline
\end{tabular}
\caption{Quantitative Comparisons of 0.1 Actual Masking Ratio (The Baseline was calculated by comparing masked text to the original text)}
\label{tab: Quantitative Comparisons of 0.1 overall}
\end{table}

\subsubsection{Hybrid Masking}
 After comparing different strategies with the same actual masking ratio, we explored hybrid masking strategies and compared them with other strategies at the same actual ratio. The results are shown in Table \ref{tab: Quantitative Comparisons for Hybrid Masking}. The first three columns have the same actual masking ratio. Masking only stopwords achieved the strongest performance among these strategies. However, when nouns are also masked along with stopwords, performance decreases, as masking nouns negatively affects the results. Despite this, it still performs better than random masking, indicating that stopwords have a greater influence than nouns. Next, we compared the last two columns. If 0.5 of nouns and 0.5 of stopwords are masked,  adding an additional 0.5 of masked verbs leads to worse performance, showing that verbs also negatively influence the model's performance.

\begin{table}[htbp]
\centering
\begin{tabular}{|c|p{2.4cm}|p{2.4cm}|p{2.5cm}|p{2.5cm}|}
\hline
\textbf{Bio\_ClinicalBERT} & \textbf{Stopwords Masking (0.8)} & \textbf{Random Masking (0.4)} & \textbf{Nouns (0.5) and Stopwords (0.5)} & \textbf{Nouns (0.5), Verbs (0.5), Stopwords (0.8)} \\ 
\hline
\textbf{Actual Masking Ratio} & 0.13 & 0.13 & 0.13 & 0.16 \\
\hline
\multicolumn{5}{|c|}{\textbf{ROUGE-1}} \\
\hline
Generation Performance      & 95.17 & 93.18 & 94.29 & 91.34 \\
\hline
Baseline        & 85.53 & 85.61 & 85.98 & 82.47 \\
\hline
\multicolumn{5}{|c|}{\textbf{ROUGE-2}} \\
\hline
Generation Performance      & 89.41 & 86.50 & 88.34 & 83.08 \\
\hline
Baseline        & 74.35 & 74.92 & 75.30 & 70.73 \\
\hline
\multicolumn{5}{|c|}{\textbf{ROUGE-L}} \\
\hline
Generation Performance     & 94.53 & 92.71 & 93.80 & 90.50 \\
\hline
Baseline        & 85.53 & 85.61 & 85.98 & 82.47 \\
\hline
\multicolumn{5}{|c|}{\textbf{BERTScore}} \\
\hline
Generation Performance     & 0.93 & 0.90 & 0.91 & 0.87 \\
\hline
Baseline        & 0.65 & 0.63 & 0.65 & 0.57 \\
\hline
\multicolumn{5}{|c|}{\textbf{Invalid Prediction Rate}} \\
\hline
Generation Performance     & 0.22 & 0.28 & 0.28 & 0.31 \\
\hline
\end{tabular}
\caption{Quantitative Comparisons for Hybrid Masking (The Baseline was calculated by comparing masked text to the original text)}
\label{tab: Quantitative Comparisons for Hybrid Masking}
\end{table}

\subsubsection{Comparison with and without (w/o) Entity Preservation}
 To further explore whether keeping entities is useful for our task, we compared our results with a baseline that does not retain any entities. The baseline was trained 
 with four epochs of fine-tuning on our dataset. Specifically, 0.4 of nouns from all tokens were randomly masked during the baseline training. In contrast, in our experiments, only eligible tokens—excluding clinical information—were selected for masking. The comparisons are shown in Table \ref{tab: Comparison with and without Entity Preservation}.

\begin{table}[htbp]
\centering
\begin{tabular}{|c|p{3.5cm}|p{3.5cm}|p{3.5cm}|}
\hline
\textbf{Bio\_ClinicalBERT} & \textbf{With Entity Preservation} (0.4 Nouns Masking) & \textbf{With Entity Preservation} (0.3 Random Masking) & \textbf{Without Entity Preservation} (0.4 Nouns Masking) \\ 
\hline
\textbf{ROUGE-1} & 97.62 & 95.10 & 97.31 \\
\hline
\textbf{ROUGE-2} & 95.12 & 90.17 & 94.46 \\
\hline
\textbf{ROUGE-L} & 97.56 & 94.86 & 93.71 \\
\hline
\textbf{BERTScore} & 0.96 & 0.93 & 0.91 \\
\hline
\end{tabular}
\caption{Comparison with and without Entity Preservation Using Bio\_ClinicalBERT}
\label{tab: Comparison with and without Entity Preservation}
\end{table}

As we can see, \textbf{when 0.4 nouns are masked while preserving entities, the models perform much better} than those without any entity preservation. Interestingly, when we randomly mask 0.3 while preserving entities, the model achieves lower ROUGE-1 and ROUGE-2 scores but higher ROUGE-L and BERTScores compared to models without entity preservation. This trend is consistent across different settings. It suggests that models preserving entities have less overlap with the original text, while they can retain the original narrative better. Additionally, the higher ROUGE-L score suggests that the step of \textbf{preserving document structure is indeed effective}.

These results also confirm our initial hypothesis that, for our objective — generating clinical letters that can keep the original meaning while adding some variety — \textbf{retaining entities} is much more effective than just fine-tuning the model. Moreover, this approach can effectively preserve useful information while avoiding overfitting.

    \subsection{Downstream NER Task}
  To further evaluate whether synthetic letters have the potential to replace the original raw letters, particularly in the domains of clinical research and model training, a downstream NER task was implemented. Two \textbf{spaCy NER} models were trained separately on \textbf{original raw letters and synthetic letters}.  Specifically, the synthetic letters were generated with 0.3 random masking while preserving entities.

As shown in Table \ref{tab: Comparison of Downstream NER Task}, spaCy models trained on original and synthetic letters showed \textbf{similar evaluation scores}. They even achieved F1 scores comparable to ScispaCy's score of 0.843. Therefore, the unmasked context does not significantly influence model understanding. Consequently, \textit{our synthetic letters can be used in NER tasks to replace real-world clinical letters, thereby further protecting sensitive information}.

\begin{table}[htbp]
\centering
\begin{tabularx}{\textwidth}{|X|X|X|X|}
\hline
\textbf{Metric} & \textbf{spaCy Trained on Original Letters} & \textbf{spaCy Trained on Synthetic Letters} & \textbf{Performance Delta ($\Delta$) } \\ 
\hline
\textbf{F1 Score} & 0.855 & 0.853 & -0.002 \\
\hline
\textbf{Precision} & 0.865 & 0.863 & -0.002 \\
\hline
\textbf{Recall} & 0.846 & 0.843 & -0.003 \\
\hline
\end{tabularx}
\caption{Comparisons on Downstream NER Task}
\label{tab: Comparison of Downstream NER Task}
\end{table}

 \subsection{Post-Process Results}
 \label{sec: Post-Process Results}
\subsubsection{Filling in the Blanks}
\begin{figure}[htbp]
  \centering
  \includegraphics[width=0.8\linewidth]{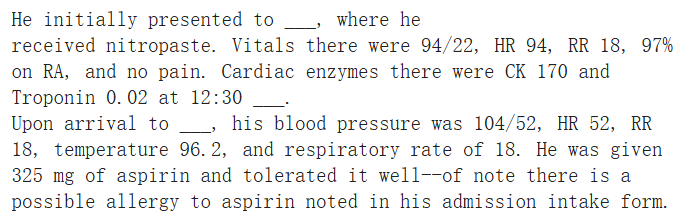}
  \caption{Example Sentence Before Post-Processing \cite{goldberger2000physiobank,johnson2024mimiciv,johnson2023mimic} (`note\_id': `16441224-DS-19')}           

  \label{fig: Example Sentence Before Post-Process}  
\end{figure}

\begin{figure}[htbp]
  \centering
  \includegraphics[width=0.8\linewidth]{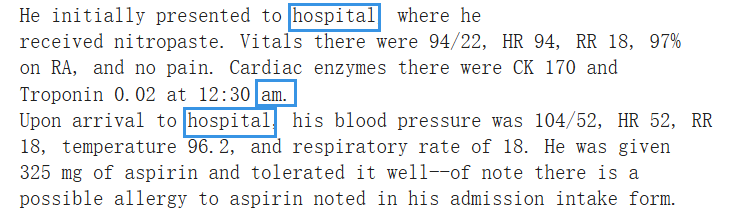}
  \caption{Post-Processing Results with BERT-Base}           

  \label{fig: Post-Process Results with BERT-Base}  
\end{figure}

\begin{figure}[htbp]
  \centering
  \includegraphics[width=0.8\linewidth]{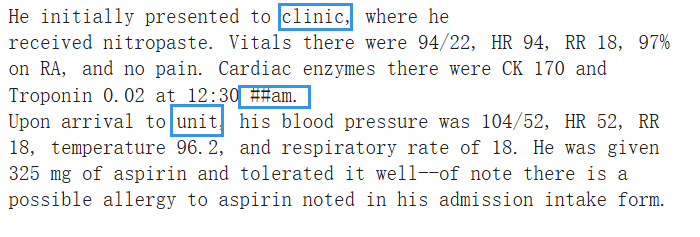}
  \caption{Post-Processing Results with Bio\_ClinicalBERT}           

  \label{fig: Post-Process Results with BioBERT}  
\end{figure}

 One example text without post-processing is shown in Fig \ref{fig: Example Sentence Before Post-Process}. After filling in the blanks, the results with BERT-base and Bio\_ClinicalBERT are shown in Fig \ref{fig: Post-Process Results with BERT-Base} and Fig \ref{fig: Post-Process Results with BioBERT} separately. We can see that both models can partially achieve the goal of making the text more complete. However, neither of them created a coherent story to fill in these blanks. They just used general terms like ``hospital" and ``clinic". Perhaps other decoder-only models, more suitable for generating stories like GPT, could perform better and should be explored in the future.

\subsubsection{Spelling Correction}
\begin{figure}[htbp]
  \centering
  \includegraphics[width=0.8\linewidth]{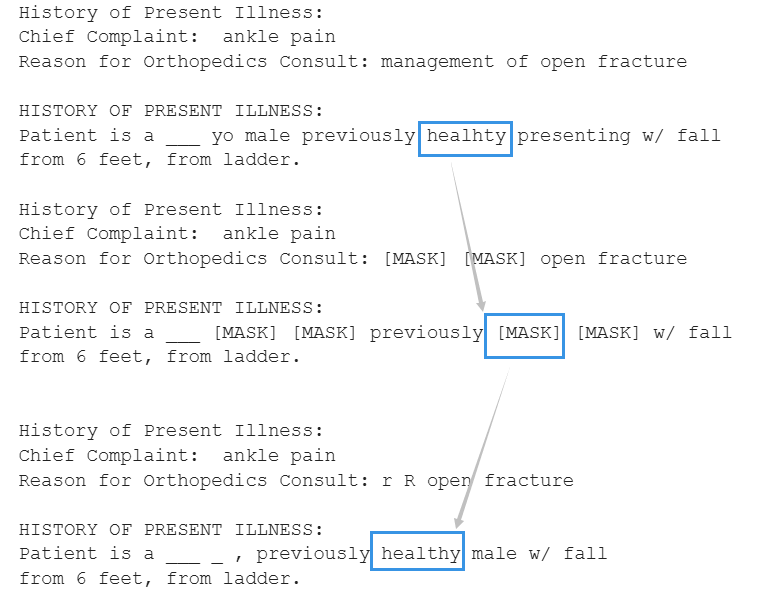}
  \caption{Spelling Correction by masking and Generating \cite{goldberger2000physiobank,johnson2024mimiciv,johnson2023mimic} (`note\_id': `10807423-DS-19')}           

  \label{fig: Spelling Correction by masking and Generating}  
\end{figure}

 Fig \ref{fig: Spelling Correction by masking and Generating} shows that if the incorrect words are masked, the models may be able to correct the misspelled tokens by predicting them. However, the masking process is random. Additionally, sometimes the predicted words will be incorrect because some models tokenize the sentence into word-pieces. Therefore, a post-processing step is necessary for correcting spelling.

\begin{figure}[htbp]
  \centering
  \includegraphics[width=0.8\linewidth]{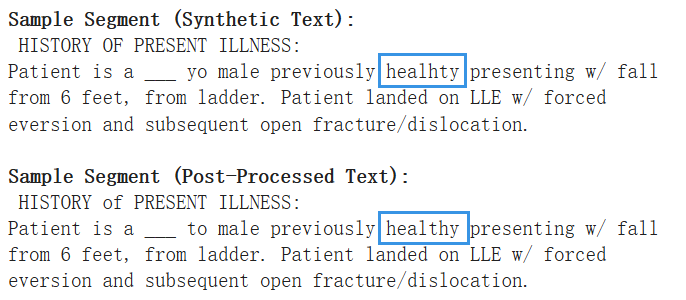}
  \caption{Spelling Correction \cite{goldberger2000physiobank,johnson2024mimiciv,johnson2023mimic} (`note\_id': `10807423-DS-19')}           

  \label{fig: Spelling Correction}  
\end{figure}

As shown in Fig \ref{fig: Spelling Correction}, Tooltik `TextBlob' \cite{textblob} can successfully correct misspelled words (`healhty') in our sample text. However, if clinical entities are not preserved during the pre-processing step, `TextBlob' \cite{textblob} may misidentify some clinical terms as spelling errors. It may be because `TextBlob' \cite{textblob} was developed on the general corpus, not a clinical one. Additionally, its corrections are limited to the word level and do not consider any context. Therefore, if words are misspelled deliberately, they could be processed incorrectly. Thus, \textit{developing a clinical misspelling correction toolkit is a promising} research direction in the future.

\section{Conclusions and Future Work}
    \label{sec: Discussion and Future Work}
    \subsection{Key Findings}
 These results provided some useful findings in generating clinical letters, including: 

\begin{itemize}
    \item \textbf{Encoder-only models generally perform much better} in clinical-letter masking and generation tasks, which is consistent with a very recent study by \newcite{DBLP:journals/corr/abs-2405-12630}. When clinical information is preserved, \textbf{base encoder-only models perform comparably to clinical-related models}.
    \item To generate clinical letters that preserve clinical narrative while adding variety, \textbf{BERTScore should be the primary evaluation metric}, with other metrics serving as supporting references. This is because BERTScore focuses more on semantic rather than literal similarity, and it is consistent with qualitative assessment results.
    \item \textbf{Different types of masked tokens influence} the quality of synthetic clinical letters. \textbf{Stopwords} have a \textbf{positive} impact, while \textbf{nouns and verbs} have \textbf{negative} impacts.
    \item For our objective, \textbf{preserving useful tokens} is more effective than just fine-tuning the model without preserving any entities.
    \item The \textbf{unmasked context} does not significantly influence the model's understanding. As a result, the \textbf{synthetic letters can be effectively used in downstream NER task} to replace original real-world letters.
\end{itemize}

    \subsection{Limitations}
 Although the strategies mentioned above help generate diverse, de-identified synthetic clinical letters, there are still some limitations in applying this method generally.
\begin{itemize}
\item \textbf{Challenges in the Dataset: } Since these clinical letters are derived from the real world, some problems are inevitable. For example, there may be spelling errors in the dataset, such as in note\_id ``10807423-DS-19", `healthy' is misspelled as `healhty'. These errors may hurt the usability of the synthetic text. In addition, some polysemous words may also cause ambiguity due to context. For example, the word `back' can refer to an anatomical entity (such as the back), and can also be used as an adverb. 
\item \textbf{Data Volume: }Due to the difficulty in collecting annotated data, only 204 clinical letters are used in our investigation. They may not be representative enough, which could limit the applicability of our findings to a broader scenario. Additionally, the data we used were already de-identified. Although we considered de-identification and took steps to mask all privacy information, the effectiveness of these approaches cannot be fully tested because we do not have access to sensitive datasets. 

\item \textbf{Evaluation Metrics: }In this project, we primarily use BERTScore for evaluation, while also incorporating other metrics such as ROUGE and readability metrics. Currently, there is \textbf{no evaluation framework that can assess all aspects }simultaneously, including maintaining the original meaning, diversity, readability, and even clinical soundness. 

\item \textbf{Clinical Knowledge Understanding: }While the model can often preserve clinical entities and generate contextually reasonable tokens, it sometimes encounters comprehension errors. For example, in a context where `LLE' (`left lower extremity') is used, Bio\_ClinicalBERT incorrectly predicted the nearby masked token as `R ankle' (`right ankle'). In this case, the model fails to capture the side knowledge accurately.

\item \textbf{Computing Resources: }Due to the limitations of computing resources, we explored a limited range of language generation models. There may be other models, such as improved decoder-only models, that could better adapt to our task.
\end{itemize}

    \subsection{Future Work}
 Based on the limitations mentioned above, we outlined some potential directions to further explore:
\begin{itemize}

\item \textbf{Test on More Clinical Datasets: }To further evaluate the effectiveness of these masking strategies, more annotated clinical letters should be tested to assess system generalisation. 
\item \textbf{Assess de-identification Performance: } A quantitative metric for de-identification evaluation should be included in the future. Non-anonymous synthetic datasets can be used to evaluate the de-identification process,  so that this system can be applied directly to real-world clinical letters in the future.
\item \textbf{Evaluation Benchmark: }A new metric that is suitable for our task should be developed. Specifically, this metric should consider both similarity and diversity. Weighting parameters for each dimension could be useful and can be obtained through neural networks.  For evaluating clinical soundness, it may be necessary to invite clinicians to assess the synthetic letters based on multiple dimensions \cite{ellershaw2024automated}. Furthermore, the mapping from clinical letters to their quality scores can be learned using deep learning.

\item \textbf{Balancing Knowledge from Both Clinical and General Domains: }Although there are numerous clinical-related encoder-only models, only a few can effectively integrate clinical and general knowledge. \newcite{xie2024me} demonstrated that mixing the clinical dataset with the general dataset in a certain proportion can help the model better understand clinical knowledge. Therefore, a new BERT-based model should be trained from scratch using both clinical and general domain datasets.
\item \textbf{Synonymous Substitution: } We focused on exploring the range of eligible tokens for masking. Additionally, a masking strategy similar to BERT's can be integrated with our results \cite{devlin-etal-2019-bert}. Specifically, we can select certain tokens to mask, some to retain, and replace others with synonyms. This approach can further enhance the variety of synthetic clinical letters. Moreover, the retained clinical entities can also be substituted using the entity linking to SNOMED CT.

\item \textbf{Spelling Correction: }As mentioned in Section \ref{sec: Post-Process Results}, there are very few toolkits available for spelling correction in the clinical domain. Standard spelling correction tools may misidentify clinical terms as misspelled words. Therefore, it is necessary to develop a specialised spell-checking tool adapted to the clinical domain.
\end{itemize}




\section*{Ethics Consideration}
There are no ethical concerns in this project. Before accessing the dataset, we completed the necessary training (CITI Data or Specimens Only Research) and signed the Data Use Agreement (DUA). In addition, the results of this project will not be used for any commercial purpose.

\section*{Acknowledgements}
We thank the insightful feedback from Nicolo Micheletti which is very valuable to the start of this work.
LH, WDP, and GN are grateful for the support from the grant “Assembling the Data Jigsaw: Powering Robust Research on the
Causes, Determinants and Outcomes of MSK Disease.” The project has been funded by the Nuffield
Foundation, but the views expressed are those of the authors and not necessarily of the Foundation. 
Visit www.nuffieldfoundation.org. 
LH, WDP, and GN are also supported by the grant “Integrating hospital outpatient letters into the healthcare data space” (EP/V047949/1; funder: UKRI/EPSRC).


\bibliographystyle{coling}
\bibliography{Frontiers2021}

\section*{Appendix}

\end{document}